\documentclass[oneside]{article}

\usepackage{PRIMEarxiv}
\usepackage{amsmath}
\usepackage[utf8]{inputenc} % allow utf-8 input
\usepackage[T1]{fontenc}    % use 8-bit T1 fonts
\usepackage{hyperref}       % hyperlinks
\usepackage{url}            % simple URL typesetting
\usepackage{booktabs}       % professional-quality tables
\usepackage{amsfonts}       % blackboard math symbols
\usepackage{nicefrac}       % compact symbols for 1/2, etc.
\usepackage{microtype}      % microtypography
\usepackage{lipsum}

\usepackage{fancyhdr}       % header
\usepackage{graphicx}       % graphics
\graphicspath{{media/}}     % organize your images and other figures under media/ folder
\usepackage{multirow}
\usepackage{bigstrut}
\usepackage{tabularx,booktabs}
\usepackage{algorithm}
\usepackage{amsmath}
\usepackage{adjustbox}
\usepackage{amssymb}
\usepackage{anyfontsize}
\usepackage{blindtext}
\usepackage{multirow}
\usepackage{amssymb}% http://ctan.org/pkg/amssymb
\usepackage{pifont}% http://ctan.org/pkg/pifont
\usepackage[dvipsnames]{xcolor}
\usepackage[dvipsnames]{xcolor}
\usepackage[labelformat=simple]{subcaption}

\DeclareCaptionLabelFormat{subcaptionlabel}{\normalfont(\textbf{#2}\normalfont)}
\usepackage{tabularx,booktabs}
\usepackage{algpseudocode} 
\usepackage{booktabs} % For formal tables
\usepackage[scr=rsfs]{mathalpha}
\usepackage{pdfpages}
\usepackage{pgfplots}
\usepackage{pgfplotstable}
\pgfplotsset{compat=1.7}
\usepackage{tikz}
\usepackage{lipsum}
\usepackage{verbatim}
\usepackage[resetlabels]{multibib}
\usepackage{adjustbox}

\usepackage{filecontents}
\usepackage{lipsum}

\DeclareMathOperator*{\concat}{%
    \mathchoice%
        {\Big\Vert}%
        {\big\Vert}%
        {\Vert}%
        {\Vert}%
}

\def\BibTeX{{\rm B\kern-.05em{\sc i\kern-.025em b}\kern-.08em
    T\kern-.1667em\lower.7ex\hbox{E}\kern-.125emX}}
\title{
Companion Animal Disease Diagnostics based on Literal-aware Medical Knowledge Graph Representation Learning}

\author{
 Van Thuy Hoang${}^{\star}$\\
  Dept. of Artificial Intelligence\\
  The Catholic University of Korea \\
  Bucheon, Republic of Korea \\
  \texttt{hoangvanthuy90@catholic.ac.kr} \\
  \And
  Sang Thanh Nguyen${}^{\star}$\\
  Dept. of Artificial Intelligence\\
  The Catholic University of Korea \\
  Bucheon, Republic of Korea \\
  \texttt{sangnguyen@catholic.ac.kr} \\
  \And
 Sangmyeong Lee \\
  Dept. of Physics\\
  The Catholic University of Korea \\
  Bucheon, Republic of Korea \\
  \texttt{sngmng@hanyang.ac.kr} \\
  \And
   Jooho Lee \\
  School of Computer Science and IE\\
  The Catholic University of Korea \\
  Bucheon, Republic of Korea \\
  \texttt{jooho0223@catholic.ac.kr} \\
  \And
  Luong Vuong Nguyen \\
  Dept. of Artificial Intelligence\\
  FPT University\\
  Danang, Vietnam \\
  \texttt{vuongnl3@fe.edu.vn} \\
  \And
  O-Joun Lee ${}^{\dagger}$ \\
  Dept. of Artificial Intelligence\\
  The Catholic University of Korea \\
  Bucheon, Republic of Korea \\
  \texttt{ojlee@catholic.ac.kr} \\
}

\begin{document}
\maketitle

\begin{abstract}
Knowledge graph (KG) embedding has been used to benefit the diagnosis of animal diseases by analyzing electronic medical records (EMRs), such as notes and veterinary records.
However, learning representations to capture entities and relations with literal information in KGs is challenging as the KGs show heterogeneous properties and various types of literal information.
Meanwhile, the existing methods mostly aim to preserve graph structures surrounding target nodes without considering different types of literals, which could also carry significant information.
In this paper, we propose a knowledge graph embedding model for the efficient diagnosis of animal diseases, which could learn various types of literal information and graph structure and fuse them into unified representations, namely LiteralKG.
Specifically, we construct a knowledge graph that is built from EMRs along with literal information collected from various animal hospitals.
We then fuse different types of entities and node feature information into unified vector representations through gate networks.
Finally, we propose a self-supervised learning task to learn graph structure in pretext tasks and then towards various downstream tasks.
Experimental results on link prediction tasks demonstrate that our model outperforms the baselines that consist of state-of-the-art models.
The source code is available at \url{https://github.com/NSLab-CUK/LiteralKG}.

\let\thefootnote\relax
\footnotetext{${}^{\star}$ These authors have equally contributed to this study as co-first authors} 
\footnotetext{${}^{\dagger}$ Correspondence: \texttt{ojlee@catholic.ac.kr}; Tel.: +82-2-2164-5516} 
%%%%%%%%%%
\end{abstract}

% keywords can be removed 

\section{Introduction}\label{sect:intro}

Identifying animal diseases early is important to prevent and control further companion animal diseases and spread.
For the diagnosis of animal diseases, pet owners mostly rely on professional veterinarians who possess general medical knowledge.
However, the lack of high-level experts and timelines could not be guaranteed, resulting in a great financial loss.
Therefore, there is a critical need to find efficient methods to assist experts in efficiently diagnosing animal diseases.
Furthermore, sick animals provide historical diagnosis records from different sources, which could bring valuable information for expert systems to explore latent knowledge.  

%Since medical applications have received much priority from the community, it is essential to construct models effectively leveraging medical data (e.g., patient information and symptoms stored in electronic medical records (EMRs), ultrasound images, x-ray images, etc.).
% To meet this requirement, existing models concentrate on representing medical data (e.g., patient information and symptoms stored in electronic medical records (EMRs), ultrasound images, x-ray images, etc.) in an effective structure that can be leveraged to diagnose correct diseases \cite{king_2018_behavior, li_2019_low, yamaguchi_2020_deep, yanaeva_2021_application, gong_2021}. 
Recently, knowledge graphs (KGs) have shown the power to solve important tasks in the biomedical area, especially in animal disease diagnostics.
KGs could represent natural relations of entities from electronic medical records (EMRs), which then be used to explore useful knowledge \cite{zhang_2019_vettag, boland2019applied, inacio_2020_automated, gong_2021, gao_2021_disease, paynter_2021_veterinary, li_2022_plaguekg, murali_2023_towards}.
%where these entities have implicit and complex constraints, i.e., some diseases are only found in a breed or species, an animal can have more than one types of diseases \cite{zhang_2019_vettag, boland2019applied, inacio_2020_automated, gong_2021, gao_2021_disease, paynter_2021_veterinary, li_2022_plaguekg, murali_2023_towards}.
Several object information in EMRs, i.e., edge, gender, and treatments, are defined as entities, and edges denote the relations of entities.
By using KGs to represent entities and their relations, learning representations of KGs could assist as an auxiliary role in providing expert decision support.
Accordingly, various graph embedding methods have been proposed to learn various types of relations among entities in an MKG \cite{li_2020_method, gong_2021}.
% In particular, many fields in a medical record can be represented as entities in the knowledge graph. Moreover, the relations among these entities are different and should be represented in different types suitable with the characteristics of a KG. 
The learned representations are then utilized to solve downstream tasks, such as disease diagnosis and medicine recommendation. \cite{gong_2021, lin_2020_kgnn, hong_2022_lagat}. 

%Furthermore, entities created from EMRs contain literals. This information refers to a specific value or data type associated with attributes (e.g., the numerical value in weight, age, or text in disease symptoms, etc.) \cite{gesese_2019_comprehensive, gesese_2021_survey}. Various graph embedding models leverage it to improve the representation of entities \cite{gesese_2019_comprehensive, gesese_2021_survey, wu_2018_knowledge, kristiadi_2019}.
% For example, entities are not linked in KG, but their attributes are similar. Their representations can be improved by leveraging.
%Moreover, learning similarity among literals helps predict the relations among entities that are not linked in the graph \cite{gesese_2021_survey}.

Most of the graph-based methods have been proposed to learn the graph structure in diagnosing animal diseases \cite{nusai2015swine, nugroho2017mobile}.
%Several existing KG-based methods mainly focus on particular areas, such as drug discovery and protein interactions \cite{wishart2018drugbank, mohamed2020discovering}.
%Otherwise, most existing KG-based models mainly exploit the graph structure information, i.e., entities and relations.
Since knowledge graphs have heterophily properties and different types of relations, the learned representations thus could be limited to capturing the implicit and complex relations in KGs.
%along with entity features and structural information, such as numerical and textual information.
For example, literal information in EMRs, i.e. symptoms, text description, and doctor’s advice for the treatments, could also benefit the models to predict animal diseases \cite{DBLP:journals/access/Chai20, DBLP:journals/bib/LanDCZLPC22, DBLP:conf/icdm/XuXSLLXW21}.
%Furthermore, the existing models mostly rely on manually pre-defined rules to learn representations which leads to the difficulty for models to expand and learn implicit and complex relationships between entities, such as symptoms and disease relations \cite{DBLP:journals/access/Chai20}.
Consequently, the performance of the models could be reduced as the models overlooked literal information.

Although there have been several studies \cite{lipton2015learning,hoangugt,DBLP:conf/icdm/XuXSLLXW21} that leverage literal information to capture the semantic relations and improve the learned representations, two limitations of the existing models restrict their ability to represent KG structure.
First, some methods treat various entity attributes and their relations equally, which could not be sufficient for learning important features of different entities.
Second, these methods mainly learn textual information through several deep learning models, i.e., RNNs and LSTM, as auxiliary modules to capture the literal information \cite{lipton2015learning}.
The entity attributes and literal information are then learned independently, which could not bring the expressiveness of the graph-based model to handle heterogeneous properties.

To address this limitation, we represent LiteralKG, a KG representation learning model that could learn both different types of literal information and graph structure and then fuse them into unified representations.
In particular, we first transform entities and their attributes into unified representations by using a gate function \cite{kristiadi_2019}. 
Since the different attributes are formed from different types of information, such as numerical and textual values, such fusion layers could benefit for representing unified vectors and then embedding models \cite{chen_2020, s23084168}.
%That vector is the initial representation of a GAT model. 
%We then encode vectors by using the GAT model and generate final representations. 
After composing literal enriched embedding vectors, the representations could be encoded and obtained from attentive embedding propagation layers \cite{velivckovic_2017,jeon2021learning}. 
Furthermore, we adopt a self-supervised learning task that could learn graph structures from pretext tasks without using any label information to generate learned representations, and then use the representations for downstream tasks, such as link prediction \cite{nguyen2023connector,Jeon2022}.
 In particular, we focus on several research questions: 
 %creating the most powerful disease diagnosis model by effectively representing entities, relations, and attributes in an MKG. To achieve this purpose, research questions are given as:
    \begin{itemize}
    \item RQ1. Could fusing different types of literal information and entity relations benefit the representation learning to discover explicit and complex relations in KGs?  
    % Due to semantic similarities among literals, leveraging this information is effective in predicting the missing relations in MKG. 
    \item RQ2. Which characteristics of encoders could benefit the model in learning various types of features and complex relations in KGs?

    \item RQ3. Could the pre-training model with pre-training tasks generate effective representations that could then benefit downstream tasks?
    
    %The triplet scoring function learns many types of relations. The pre-trained model with this function boosts the performance of downstream tasks.
    % Regarding the representation of relationship types in MKG, whether a pre-trained model that utilizes the power of the triplet loss function is highly efficient for disease diagnosis. 
    
    %\item RQ 3. Since initial residual connection and identity mapping are methods to retain information from the node feature and the graph structure across GNN layers \cite{chen_2020}, they may bring more benefits for the deep GNN model to represent entities. 
    \end{itemize}

To answer RQ1, we aim to conduct experimental cases to explore which types of attributes could benefit the learned representations and contribute to the overall performance of LiteralKG.
Since there are two types of literal features, i.e., numerical and textual information, it is critical to investigate combinations between different forms of features that can contribute to the effective performance of LiteralKG.
For RQ2, we aim to investigate the performance of our model on different types of GNN aggregators.
It is worth noting that most GNN aggregators are initially designed for homogeneous and simple graphs.
However, applying GNN aggregators to KGs could be different as KGs show heterophily properties and complex relations. 
Furthermore, we aim to investigate which aggregators could be appropriate and show a powerful aggregation function to capture the complex relations in our data. 
For RQ3, we investigate the efficiency of the pre-trained model trained by optimizing a triplet loss function.
Pre-training tasks could learn representations to capture underlying relations and structures without using labelled information. 
We aim to investigate whether learned representations could benefit downstream tasks \cite{zhang_2020_graphbert, s23084168, joo_2023_classification}. 
Therefore, the pre-training model is expected to bring more efficiency in extracting complex relations and benefit downstream tasks.

% The three above questions are considered to clarify the importance of leveraging literal information and representing many types of relations among entities. In the constructed MKG, there are two types of literal features, numerical and textual. To answer RQ 1, experimental cases are set to learn how efficient each type is. With an MKG containing many textual features, these attributes are supposed to be more efficient than numerical attributes. 
% RQ 2 assumes the efficiency of the pre-trained model trained by a triplet loss function.
% In graph representation learning, the pre-trained model can significantly improve the performance \cite{zhang_2020_graphbert, s23084168, joo_2023_classification}. Therefore, this architecture is expected to bring more efficiency in representing many types of relations and strengthen the performance of disease diagnosis.
% Since residual connection and identity mapping can deal with the over-smoothing problem, these methods are assumed to be effective in aggregating high-order neighbors, as mentioned in RQ3. As a result, these methods may be effective for learning a large-scale graph in which considering global graph structure is essential to represent entities.

% Moreover, we construct a MKG from raw data collected from 32 animal hospitals. Based on the different characteristics of EMRs, there are various experiments are conducted to identify the most efficient GNN aggregation technique that can deal with the MKG. 

%%%%%%%%%%%%%%%%%%%%%%%%%%%%%%%% HERE %%%%%%%%%%%%%%%%%%%%%%%%

The contributions of this paper are summarized as follows:

\begin{itemize}

  \item We construct a medical knowledge graph that comprises 595,172 entities and 16 relation types from various EMRs.
  
  \item We propose LiteralKG, a knowledge graph embedding model that could learn different types of literal information and graph structure and then fuse them into unified representations.
  
  \item  We propose a self-supervised learning task that could learn the graph structure from pretext tasks to generate representations, and then the pre-trained model is used for downstream tasks to predict animal diseases.
  
  % \item To represent the complicated relations among entities in MKG, a triplet loss function is employed in the pre-training phase. In fine-tuning, the pre-trained embedding vectors are utilized to predict animal diseases.
  
  \item 
  The experimental results on the KG with different types of GNN aggregators and residual connection and identity mapping show the superiority of LiteralKG over baselines.
  
  % \item The proposed model is compared with several baseline models. From the evaluation results, our model outperforms them by leveraging the literal information and adopting a pre-trained phase. Moreover, since there are some settings of the proposed model, including the pre-training phase, adapting residual connection, etc, we establish several strategies and conduct various experiments to evaluate the model and observe the capability of these settings.
\end{itemize}

The rest of this paper is organized as follows. 
Section \ref{section_2} presents literature reviews of existing methods to solve the above problems.
Section \ref{section_3} describes our proposed strategies to construct a medical KG from EMRs. 
The methodology of LiteralKG is presented in section \ref{section_4}. 
Section \ref{section_5} shows the experimental results and analysis.
Section \ref{section_6} is the conclusion and future work.

\section{Related work}
\label{section_2}

Over recent years, several graph-based methods have been proposed to handle medical records through learning graph structure \cite{zhao_2018_emr, gong_2021, li_2020_method, DBLP:conf/ijcai/LeeJ20,DBLP:journals/ai/LeeJ20,DBLP:conf/ijcai/ParkK20}. 
For example, several studies \cite{DBLP:journals/access/Chai20, DBLP:journals/bib/LanDCZLPC22} use translation-based methods, such as TransE \cite{bordes_2013} and TransR \cite{lin_2015}, to map entities and relations into latent space and then predict diseases.
PrTransX \cite{li_2020_method} enhances translation-based methods, such as TransE \cite{bordes_2013}, TransH \cite{wang_2014_knowledge}, TransR \cite{lin_2015}, TransD \cite{ji_2015_knowledge}, or TranSparse \cite{ji_2016_knowledge} by optimizing triplet probability into a scoring function and the margin-based loss function to learn representations.
Gong et al. \cite{gong_2021} have proposed a model representing diseases, medicines, and patient data in EMRs by utilizing a KG triplet loss function. 
However, most existing models aim to learn entity representations without considering the literal information, i.e. symptoms and doctor's advice, which could carry significant information for learning representations \cite{dettmers_2018_conv2DKG,DBLP:journals/fdata/JeonLJ19}. 
In contrast, our model could learn different side information types, such as numeric and literal features.
Several studies have been proposed to capture entity features and literal information to enrich learned representations \cite{DBLP:conf/ijcai/LeeJ20}.
For example, Tay et al. \cite{tay_2017_multi} introduce AttrNet model to learn the entities and their relations in triplets through the combination with attribute features.
Wu et al. \cite{wu_2018_knowledge} have proposed TransEA, which can represent the numerical features of entities and learn the attribute triplet based on an attribute score. 
Kristiadi et al. \cite{kristiadi_2019} have proposed the LiteralE model to learn the numerical and textual features through linear transformation and optimize a triplet scoring function.
Li et al. \cite{li_2022_fusing} have utilized the bilinear feature multiplication in a multi-model fusion to learn the text and image attributes with the entities in the KGs. 
However, most existing models learn representations through linear transformations, which could suffer from slow convergence and the capability to capture the heterogeneous property and graph structures in KGs.

Several studies have been proposed to automatically diagnose animal diseases through constructing an MKG and then learning the entities and relations \cite{yanaeva_2021_application, yamaguchi_2020_deep, DBLP:journals/fdata/JeonLJ19}.
% A study based on support vector machine (SVM) \cite{wan_2010_animal} uses a hyperplane to classify the correct animal disease in MKG. 
% An RNN-based model has been proposed to predict animal disease by mining graph data constructed from EHRs \cite{kokkinos_2022_early}.
% These methods are applied in predicting animal diseases that are limited in capturing graph structure and modelling many types of relations which can be solved in our approach.
%In particular, shallow KG embedding models or deep models can capture the graph structure to predict the relations of missing entities with the existing entities \cite{gong_2021, lin_2020_kgnn, hong_2022_lagat}. 
For example, to diagnose dairy cows' diseases, Gao et al. \cite{gao_2021_disease} have constructed an MKG from EMRs and learned representations by using TransD method.  
Several studies adopt GAT to represent entities and relations in KGs and then combine them with the RNNs model to capture the patient’s history for predicting disease \cite{DBLP:journals/access/Chai20, DBLP:conf/icdm/XuXSLLXW21}. 
For example, Xu et al. \cite{DBLP:conf/icdm/XuXSLLXW21} use GAT to learn representations through the combination with LSTMs as an auxiliary module for diagnosing pathology and disease \cite{DBLP:journals/access/Chai20}. 
%Xu et al. \cite{DBLP:conf/icdm/XuXSLLXW21} aim to constructed a framework containing a pre-trained model representing entities in MKG using GAT. The pre-trained embeddings are then trained in an RNN model to capture the patient’s history for predicting disease.
However, learning KG structures independently with literal information may not enrich the learned representations and eventually reduce the model performance \cite{gesese_2021_survey}.
Unlike existing models, our model could learn different types of literal information combined with graph structures. 
Furthermore, the existing models learn different types of attributes with uniform weights between different entities and may not capture important "messages".
In contrast, our model first fuses different types of entities and literal information through gate networks.
Then, LiteralKG learns vector representations through coefficients between triplets in the KG, which could benefit from capturing graph structure using attentive weights.

\section{Medical Knowledge Graph Construction with Electronic Medical Records}
\label{section_3}

\begin{table*}
\caption{
Summary of entities, relations, and attributes in our MKG.
There are various types of relations between entities, including one-to-one, one-to-many, and many-to-many. 
Several entities, e.g., age, weight, and disease, carry the numerical or textual attributes described in the Attribute column.}
\label{tab:2}
    % \newcolumntype{D}{>{\centering\arraybackslash}X}
    % \begin{tabularx}{\fulllength}{p{1.5cm}p{1.5cm}p{1.3cm}p{7cm}p{1.5cm}p{3.5cm}}
    \begin{tabularx}{\textwidth}{p{0.08\textwidth}p{0.06\textwidth}p{0.06\textwidth}p{0.3\textwidth}p{0.15\textwidth}p{0.2\textwidth}}
        \toprule
        \textbf{Entity}	& \textbf{\#items} & \textbf{Notation} & \textbf{Description} & \textbf{Relation} & \textbf{Attribute}\\
        \midrule
			
Medical Record & 86,537 & $\mathbb{M} $ & The medical records of the visited companion animals that are connected to entities with various information about the animal, such as symptoms, age, weight, prescription, and the veterinarian's opinion of the companion animal who came for a checkup.
 & 
$r_A: \mathbb{M} \rightarrow \mathbb{A}$

$r_D: \mathbb{M} \rightarrow \mathbb{D}$

$r_Y: \mathbb{M} \rightarrow \mathbb{Y}$

$r_P: \mathbb{M} \rightarrow \mathbb{P}$

$r_T: \mathbb{M} \rightarrow \mathbb{T}$

$r_C: \mathbb{M} \rightarrow \mathbb{C}$

$r_E: \mathbb{M} \rightarrow \mathbb{E}$

$r_U: \mathbb{M} \rightarrow \mathbb{U}$

$r_W: \mathbb{M} \rightarrow \mathbb{W}$

$r_G: \mathbb{M} \rightarrow \mathbb{G}$
 & -\\
 
\hline
Animal & 12,545 & $ \mathbb{A} $ & It has a corresponding one-to-one number to the visited companion animal and is connected with the information with breed and species, and gender.

e.g., "201-5010664"
 & 
 $r_B: \mathbb{A} \rightarrow \mathbb{B}$
 
 $r_S: \mathbb{A} \rightarrow \mathbb{S}$

&-\\

\hline

Species & 23 & $ \mathbb{S} $ & The species of the visited companion animal.

e.g., "Canine" &
 - & -\\ 
 \hline

Breed & 607 & $ \mathbb{B} $ & The breed of the visited companion animal.

e.g., "Poodle’"
 & - & -\\ 
 \hline
 
%\bottomrule
%		\end{tabularx}
%	\end{adjustwidth}
%\end{table}

%\begin{table}[H]\ContinuedFloat
%\centering \small
%\caption{{\em Cont.}}
%\begin{adjustwidth}{-\extralength}{0cm}
%		\begin{tabularx}{\fulllength}{p{1.5cm}p{1.5cm}p{1.3cm}p{7cm}p{1.5cm}p{3.5cm}}
%        \toprule
%        \textbf{Entity}	& \textbf{\#items} & \textbf{Notation} & \textbf{Description} & \textbf{Relation} & \textbf{Attribute}\\
%        \midrule
%
 Disease & 133 & $ \mathbb{D} $ & It means a disease in which a visited companion animal has or has been infected. The disease can be linked to the “Disease Category” entity. &  $r_I: \mathbb{D} \rightarrow \mathbb{I}$& Textual attribute
 
e.g., "liver tumor"
\\ 
 \hline
 Symptom & 86,537 & $ \mathbb{Y} $ & The symptom of the visited companion animal at that time. & - & Textual attribute in the form of a sentence or word
 
e.g., "Vomiting"
\\ 
 \hline
Drugs & 1,563 & $ \mathbb{R} $ & The prescribed drug for a companion animal. 
e.g., "ANB065"
 & - & -\\ 
 \hline
Prescription & 36,791 & $ \mathbb{P} $ & The prescription such as the drug dose.
This is more detailed than “Drugs” entity.
 & $r_R: \mathbb{P} \rightarrow  \mathbb{R}$ & Textual attribute 
 
 e.g., "1 time 250mg, 3 times a day Amoxicillin/clavulanic acid Tab."\\ 
 \hline
Treatment Code & 5,952 & $ \mathbb{O} $ & There are codes corresponding to treatment. 

e.g., "A022"
 & - & - \\ 
 \hline
 Treatment & 146,113 & $ \mathbb{T} $ & The companion animal can be received a treatment plan in the hospital. The treatment can be signed as a code. In this case, it is connected to “Treatment Code” entity & $r_O: \mathbb{T} \rightarrow \mathbb{O}$ & Textual attribute

 e.g., "Intravenous injection"
 \\ 
 \hline
Comment & 132,926 & $ \mathbb{C} $ & It is described by the veterinarian about the treatment response and disease progression. & - & Textual attribute

e.g., ‘‘The animal breaths uncomfortably”
\\ 
 \hline
 Age & 24 & $ \mathbb{E} $ & Natural numbers meaning the age of the visited companion animal are put in this entity’s attribute. & - & Numerical attribute

e.g., 14
\\ 
\bottomrule
\end{tabularx}
\end{table*}

\begin{table*}\ContinuedFloat

\caption{{\em Cont.}}
 \begin{tabularx}{\textwidth}{p{0.08\textwidth}p{0.06\textwidth}p{0.06\textwidth}p{0.3\textwidth}p{0.15\textwidth}p{0.2\textwidth}}
        \toprule
        \textbf{Entity}	& \textbf{\#items} & \textbf{Notation} & \textbf{Description} & \textbf{Relation} & \textbf{Attribute}\\
        \midrule

 Age Group & 4 & $\mathbb{U} $ & There are four types of age groups including 
 "Infancy"(age < 1), "Adult" (1 $\leq$ age < 7), "Old-age" (7 $\leq$ age < 13), "Super-aged" (13 $\leq$ age)
 & - & -\\ 
 \hline
 Gender & 4 & $\mathbb{G} $ & There are only four genders of companion animals in our MKG, including unspayed female, spayed female, unneutered male, and neutered male. & - & -\\ 
 \hline
 Weight  & 85,397 & $ \mathbb{W} $ & The weight of the visited companion animal.  & - & Numerical attribute

e.g., 2.5 
\\ 
 \hline
Disease Category & 16 & $\mathbb{I}$ &  It is the category of disease. Different diseases can be connected to the same disease category.

e.g., "nervous system"
& - 
 & -\\
\bottomrule
\end{tabularx}

\end{table*}
%%%%%%%%%%%%%%%%%%%%%%%%%%%%%%%%%%%%%%%%%% 

\begin{figure}[t]
    \centering
    \includegraphics[width = 0.7\textwidth]{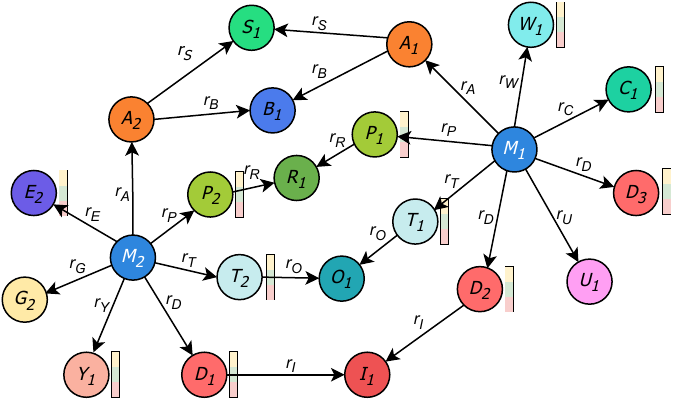}
    \caption{An example of a sub-graph from our knowledge graph.
    The circles and the rectangles surrounding entities denote entities and their attributes, respectively. 
    Several entities, such as $M$, $A$, $R$, $O$, and $D$, denote the Medical records, Animals, Drugs, Treatment code,  and Disease, respectively.} 
    \label{fig:1}
\end{figure}
% - your Medical Knowledge Graph for companion animals

% -- Schema

% -- how you constructed

% - your tasks\\
% \newcolumntype{F}[1]{>{\centering\let\newline\\\arraybackslash\hspace{0pt}}m{#1}}
% \newcolumntype{R}[1]{>{\raggedRight\let\newline\\\arraybackslash\hspace{0pt}}m{#1}}

We now represent our strategy to construct a medical knowledge graph from EMRs.
A knowledge graph (KG) is a semantic network that represents heterogeneous data with different types of entities and relations in the real world \cite{bordes_2013,lin_2015}. 
Formally, a KG is a set of triples where each triple is formed of $\langle h, r, t \rangle$, where $h, r, t$ refers to the head, relation, and tail, respectively \cite{dai_2020_survey}. 
%They contain pairs of entities and the corresponding relation types. 
Medical knowledge graph (MKG) is a knowledge graph that represents the relations in the healthcare area through representing medical data, i.e., electronic medical records (EMRs) \cite{gong_2021}. 
These records contain various types of information, such as patients, diseases, medicines, and symptoms \cite{shang_2021_ehr, gong_2021}. 
%As a result, entities and their relations represent medical properties and their constraints in EMRs.
% \textbf{How to construct a Medical Knowledge Graph from Electric Medical Records?} 
% 85,965 collected electronic medical records from companion animal hospitals.

We now explain our proposed strategy to construct a KG and entity relations from EMRs.
In our study, we constructed a KG, which is composed of 85,965 EMRs from 31 companion animal hospitals, collected from IntoCNS company\footnote{\url{http://intoh.monoalliance.com/en/}}. 
In EMRs, a record is a collection of medical properties, such as companion animal, symptoms, disease, and the veterinarian's decisions for each companion animal visit. 
We generate entities from the medical properties. 
There are a total of sixteen entity types and fifteen relation types in our KG. 
Table~\ref{tab:2} shows the detailed statistics of entity types and their relations.
In each record, the types and names of the entities are extracted from the fields and elements in EMRs, respectively. 
%Moreover, entities generated from a record are connected by a relation type. 
%For example, since the information of an animal includes breed, species, etc., an animal entity generated for it is directly connected to the following breed and species entities in different relation types. 

Figure~\ref{fig:1} illustrates simplified entities and their relations in our MKG.
Our purpose is first to construct entities and then build different types of relations between entities in KGs from EMRs.
Let $\mathbb{M}$ denote the set of medical record entities, and  $\mathbb{A}$ refers to the animal entity set.
We construct one-to-many relations between animal entity $\mathbb{A}$ and $\mathbb{M}$ since an animal $A_i$ could be examined many times and have more thane one medical record.
Other entities in our MKG are constructed into a structure that satisfies their natural types of relations.
For example, the relations between $\mathbb{M}$ and $\mathbb{D}$ are many-to-many since one animal could suffer many diseases or many animals have the same diseases.
Note that the entities are composed of different types of attributes, such as numeric and text attributes. 
%Even though entities are in separate groups, their embeddings and attributes are incorporated and transformed into a shared space \cite{li_2022_fusing}. 
%This study considers the similarity among attributes to improve the embedding model. 
%While constructing the graph, attributes are used to classify entities into respective groups before embedding them.
For example, the disease, symptom, treatment, age, and weight entities have textual or numerical attributes that can be divided into numeric or text groups. 
For textual attributes, we then encode the attributes by using Fasttext \cite{bojanowski_2017_fasttext} to capture the semantic contexts of the textual attributes. 
%For example, $C$ and $T$ entities containing veterinarian judgments were embedded into the vectors. 
%Their embeddings are similar if there is a semantic similarity among them. 
%For the numerical attributes, the attribute vectors are defined as $n_i = (A_i, W_i) \in \mathbb{R}^2$. 
%They are created from their values and are normalized by the min-max algorithm. 
Otherwise, if an entity does not contain an attribute, its attribute embedding should be known as a non-attribute entity. 
In this case, the textual or numerical attribute vectors will be presented as vectors of zero. 
Accordingly, we then fuse entity and attributed features through gate networks to generate literal enriched embedding vectors for entities. 

\begin{figure*}
    \centering
    \includegraphics[width = .96\textwidth]{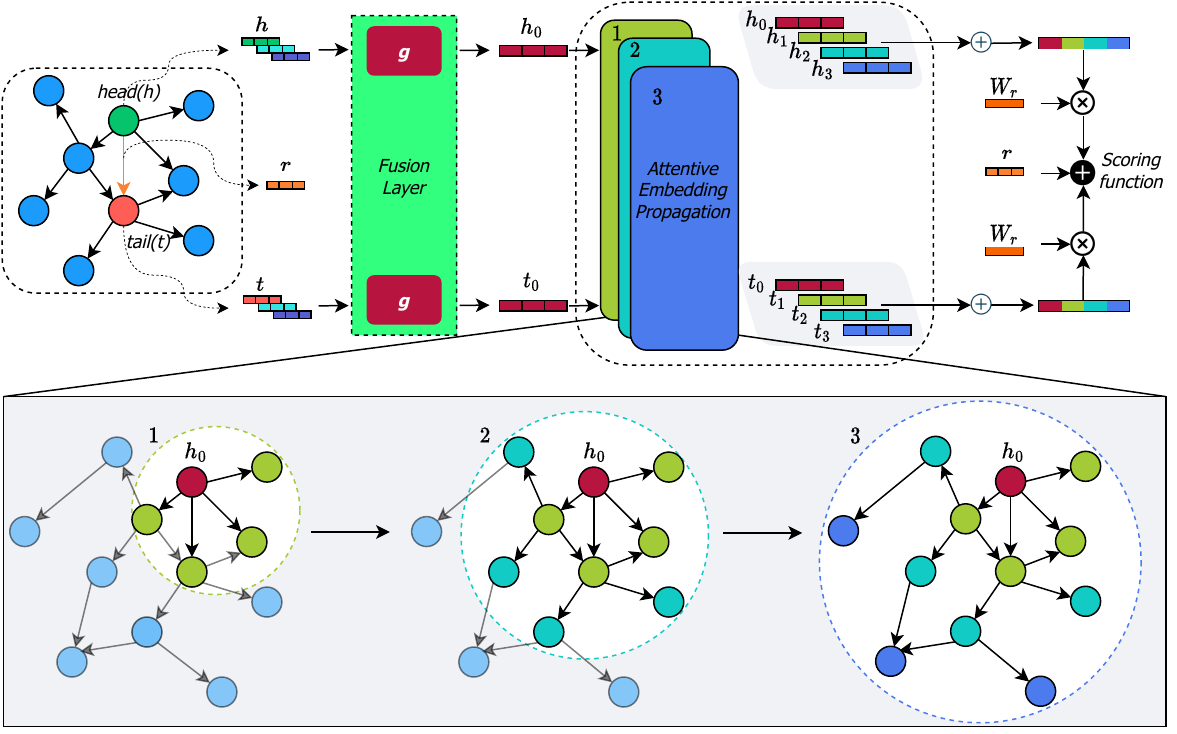}
    \caption{The overall architecture of LiteralKG. 
    The entities and attributes are fused into unified vectors through a gate function $g(\cdot)$.
    The unified vectors are then passed through attentive embedding propagation layers to generate the output representations. 
    The output representations are then concatenated with the initial embeddings to generate final representations by optimizing a score function. 
    } 
    \label{fig:3}
\end{figure*}

%%%%%%%%%%%%%%%%%%%%%%%%%%%%%%%%%%%%%%%%%%
\section{Literal-aware Medical Knowledge Graph Representation Learning}
\label{section_4}

In this section, we first represent how to fuse entities and different attribute types into unified representations. Then, we will introduce the architecture of our model in detail.
In addition, we represent a pre-training task that could learn the representations in the pretext tasks and then apply them to downstream tasks, i.e., predicting diseases.
%the techniques, including the gate mechanism and methods applied for GNN to effectively represent entities. Finally, this model can be utilized to develop real-world applications, particularly disease diagnosis.

% \subsection{Fusion Layer for representing Attributed Graph}
% \subsection{Attribute transformation in Fusion Layer}
% \subsection{Combining Attribute in Fusion Layer}
\subsection{Fusing Entity and Attribute Features}
\noindent

As mentioned earlier, the entities contain two main types of attributes, including numerical and textual attributes.
%These features help identify the relationships between entities based on semantic similarity \cite{kristiadi_2019}. 
%This advantage improves prediction accuracy when discovering the missing relations. 
%Hence, this study aims to leverage attributes. In our MKG, textual and numerical attributes can be known as feature vectors. 
%Since entity vectors and their attributed vectors are in separate spaces, they are transformed into a shared space to be learned jointly during training. 
We first design a fusion layer that contains a gate function to transform different types of attributes into unified vectors.
The textual attributes, such as disease, symptom, prescription, treatment, and comment, are transformed into vectors through Fasttext \cite{bojanowski_2017_fasttext}.
%In our KGs, two entities contain numerical attributes, including age and weight. 
Numerical attributes, such as age and weight entities, are normalized and transformed directly into feature vectors \cite{kristiadi_2019}.
In several cases, missing values are represented by `-1' as missing features.
%For example, a vector $(-1, 3)$ is generated for the numerical attribute vector of the `weight' entity with an attribute of 3kg and `-1' refers to a missing value.
%age, and weight entities carry textual or numerical attributes. 
%The former is embedded into vectors by using Fasttext \cite{bojanowski_2017_fasttext} to learn their semantic similarity. 
%The latter can be constructed into feature vectors \cite{kristiadi_2019}. 
%For entities not containing this attribute, its vector will be received -1 values in all features of the numerical vector as $<-1, -1>$. 
%its vector will be $(age, weight)$. 
%For example, a vector $(-1, 3)$ denotes the "weight" entity with an attribute of 3kg and `-1' refers to non-attributes. 
%These vectors are then normalized using min-max normalization  into the fusion layer. 
We then linearly transform entities into the shared space to generate the unified literal-enriched vectors by using a gate function \cite{kristiadi_2019}. 
Formally, the output representations for an $i$-th entity can be defined as:

\begin{align}
    &h_{i}^{(0)} = g(e_i,n_i,t_i) = \mu \odot \nu + ( 1- \mu ) \odot e_i \ ,  \\
    \text{where,  }  &\mu = \sigma_{1}\left (W_{E} \cdot e_{i} + W_{N} \cdot n_{i} + W_{T} \cdot t_{i} + b \right ) , \ \nu = \sigma_{2}\left ( W \cdot \left({e_{i} \|  n_{i} \| t_{i}}
    \right)\right) , \notag
\end{align}
and $\odot$ is the Hadamard product, $\sigma_1$ and $\sigma_2$ are sigmoid and tanh activation functions, respectively.
The entity vector $e_i \in \mathbb{R}^{E}$ , the numerical attribute vector $n_i \in \mathbb{R}^{N}$ and the textual attribute vector $t_i \in \mathbb{R}^{T}$ are combined and transformed into a vector $h_{i}^{(0)} \in \mathbb{R}^{h}$ with fixed dimension $h$ by a gate function $g$.
$W_E \in \mathbb{R}^{E \times h}$, $W_N \in \mathbb{R}^{N \times h}$,  $W_T \in \mathbb{R}^{T \times h}$ and $b \in \mathbb{R}^{h}$ and  $W \in \mathbb{R}^{(E + N + T) \times h}$ are learnable parameters.

% \subsection{How powerful leveraging the graph structure?}
% \subsection{Leveraging the whole graph structure}
\subsection{Learning Global and Local Structural Features}

After generating literal-enriched embedding vectors, we use an attention mechanism to learn the co-coefficients across triplets. 
Figure~\ref{fig:3} shows the overall architecture of our model. 
The literal-enriched embedding vectors will be passed through the attentive embedding propagation layers. 
Formally, the representation of a entity $e_i$ at $l$-th layer could be updated as:
\begin{align} 
h^{(l)}_{i} =& \text{  } f_{*} {\left(h_i^{(l-1)}, h_{\mathcal{N}_{i}}^{(l-1)} \right )} \ , \\
h_{\mathcal{N}_{i}}^{(l-1)} =& \sum_{\forall {(h_{i},r_{j},t_{k})} \in \mathcal{N}_{i}} \pi(h_{i},r_{j}, t_{k}) t_{k}^{(l-1)} \ ,
\end{align}
where $h^{(l)}_{i}$ is the output representation of an entity $e_i$ at $l$-th layer, $\mathcal{N}_{i}$ denotes a set of neighbours of entity $e_i$ which is composed with the triple $(h_{i},r_{j},t_{k})$ if and only if there is a link between $h_{i}$ and $t_{k}$, and $f_{*}$ is a GNN aggregator. 
%Since the attention mechanism shows efficiency in reducing noise data in deep models. Furthermore, this technique was adopted into GNN to leverage the variable  neighborhood messages \cite{velivckovic_2017}. In this work, neighborhood messages of an entity in GNN layers are multiplied by the corresponding attention scores of the entity triplets. 
 
Since neighbours with different relation types could contribute differently to a target entity, we aim to aggregate features of neighbouring nodes with attentional weights to benefit the model in learning different relations between entities.
The attentive weights could, therefore, describe the nature of the relations between entities in KGs.
More formally, the attentive scores are computed as follows:
\begin{align}
\pi(h,r,t) &= (W \cdot t)^{\intercal} \tanh(W \cdot h + r) \ , \\
\pi(h,r,t) &:= \frac{exp(\pi(h,r,t))}{\sum_{{(h,r^{'},t^{'})\in\mathcal{N}_{i}}}exp(\pi(h,r^{'},t^{'}))} \ .
\end{align}

%In our MKG, thousands of entities are connected and interleaved, making a large dense graph. Moreover, The constructed MKG contains many high degrees nodes, which create a star shape. 
%As such, leveraging this data structure is challenging for GNN aggregation. 
Since different GNN aggregators show individual characteristics, they could benefit in different ways to explore the explicit and complex KG structures.
Therefore, we aim to investigate which types of GNN aggregators could contribute to the overall performance of our model.
In this study, we use four aggregators, including GCN, GraphSAGE, Bi-Interaction, and GIN aggregators.
First, GCN aggregator \cite{wang_2019} combines the entity feature and its neighbour's features by a sum operator. At each $k$-th layer, the model aggregates $k$-hop neighbourhood features to leverage the graph structure and generate output representation. 
A non-linearity transformation is then utilized to transform the output features before updating the representations. 
% Furthermore, the more GCN layers are added, the higher the capability to capture global graph structure. 
The GCN aggregator is defined as:
% \begin{equation} 
% f_{G}{(h_i^{(l-1)}, h_{\mathcal{N}_{i}}^{(l-1)})} =\sigma(\mathbf{W}^{(l-1)}(h_i^{(l-1)}+h_{\mathcal{N}_{i}}^{(l-1)})) \ .
% \end{equation}
\begin{equation} 
f_{GCN}^{(l)}{\left (h_i^{(l-1)}, h_{\mathcal{N}_{i}}^{(l-1)}\right )} =\sigma\left(W^{(l)}\left(h_i^{(l-1)}+h_{\mathcal{N}_{i}}^{(l)}\right)\right) \ ,
\end{equation}
where $\sigma$ denotes the activation function, $h_i^{(l-1)}$ and $h_{\mathcal{N}_{i}}^{(l-1)}$ are the output of the previous layer and the aggregated neighbourhood features of an entity, respectively, and $W^{(l)}$ is the learnable transformation matrix at $l$-th layer. 

GraphSAGE aggregator replaces sum by concatenation operator to distinguish between the entity feature and its neighbourhood aggregation feature \cite{hamilton_2017, wang_2019}. 
Note that the difference between GCN and GraphSAGE is that GraphSAGE learns the topological structure for each target node neighbourhood through random walks, which could then calculate node embeddings in an inductive manner.
GraphSAGE aggregator is described as follows:
% \begin{equation}
% f_{S}{(h_i^{(l-1)}, h_{\mathcal{N}_{i}}^{(l-1)})} = \sigma(\mathbf{W}^{(l-1)}(h_{i}^{(l-1)}||h_{\mathcal{N}_{i}}^{(l-1)})) \ .
% \end{equation}
\begin{equation}
f_{SAGE}^{(l)}{\left (h_i^{(l-1)}, h_{\mathcal{N}_{i}}^{(l-1)}\right)} = \sigma\left(W^{(l)}\left(h_{i}^{(l-1)} \| h_{\mathcal{N}_{i}}^{(l-1)}\right)\right) \ ,
\end{equation}
where $\|$ is the concatenation.

Wang et al. \cite{wang_2019} have proposed a Bi-Interaction aggregator which combines the GCN-based strategy and the element-wise product of the target node and its neighbour feature. %Since this term creates more permutations in the feature aggregation, it improves the distinction of output entity representation at GNN layers and helps pass more messages. 
The element-wise product could assist the model in learning the similarity between target nodes and neighbourhoods.
Formally, the Bi-Interaction equation is considered as follows:
% \begin{equation} 
% f_{BI}{(h_i^{(l-1)}, h_{\mathcal{N}_{i}}^{(l-1)})} = \sigma_1(\mathbf{W}_{1}^{(l-1)}(h_{i}^{(l-1)}+h_{\mathcal{N}_{i}}^{(l-1)})) + \sigma_2(\mathbf{W}_{2}^{(l-1)}(h_{i}^{(l-1)}\odot h_{\mathcal{N}_{i}}^{(l-1)})) \ .
% \end{equation}
\begin{align} 
f_{BI}^{(l)}{\left(h_i^{(l-1)}, h_{\mathcal{N}_{i}}^{(l-1)}\right )} =\sigma\left(W_{1}^{(l)}\left(h_{i}^{(l-1)}+h_{\mathcal{N}_{i}}^{(l-1)}\right)\right) +  \sigma\left(W_{2}^{(l)}\left(h_{i}^{(l-1)}\odot h_{\mathcal{N}_{i}}^{(l-1)}\right)\right) \ \notag ,
\end{align}
where $W_{1}^{(l)}$ and $W_{2}^{(l)}$ are the learnable parameters.
%transformation matrices of sum and element-wise aggregator at layer $l$.

Inspired by 1-d Weisfeiler-Lehman (WL) isomorphism testing, GIN \cite{xu_2018} aims to maximize the GNNs power up to $1$d-WL test.
The key difference between GIN and other aggregators is that GIN could map different sub-structures into different representations, leading to the power to distinguish non-isomorphic sub-structures. 
%a dense connection in GIN \cite{xu_2018} is computed by incorporating all the features of GNN layers to consider the whole graph structure. Hence, it is more efficient in generating entity representations. 
%This concept helps generate more expressive features, which can be known as one way to deal with the over-smoothing problem. 
%In this study, a GIN aggregator is constructed from this technique. 
At each layer, the GIN aggregator updates the representations through the entity and neighbour features with a sum aggregator.
Note that GIN does not include any normalization during updating node features.
The GIN aggregator can be described as:

% \begin{equation} 
% f_{GIN}{(h_i^{(l-1)}, h_{\mathcal{N}_{i}}^{(l-1)}}) =  ((1+\epsilon^{(l)})\cdot h_{i}^{(l-1)} +  h_{\mathcal{N}_{i}}^{(l-1)})) \ .
% \end{equation}
\begin{align} 
f_{GIN}^{(l)}{\left (h_i^{(l-1)}, h_{\mathcal{N}_{i}}^{(l-1)}\right )} &= \sigma\left[FC\left(\left(1+\epsilon^{(l)}\right)\cdot h_{i}^{(l-1)} +  h_{\mathcal{N}_{i}}^{(l-1)}\right)\right], \nonumber \\
    %h^{(l)}_{i} &= \sigma\left(W^{(l)} \cdot \sum_{k=0}^l \left(h_{i}^{(k)}\right)\right),
\end{align}
where $\epsilon$ can be a fixed scale or learnable parameter, and $FC$ denotes the fully connected layer.

%In this study, we evaluate the proposed model on different numbers of GNN layers. 
%Although message-passing mechanism helps the model to leverage the graph structure, it can lead to an over-smoothing problem when combined with smoothed features generated from depth GNN \cite{zhao_2019_pairnorm}. 
Stacking more GNN layers can lead to the over-smoothing problem, and eventually lead to reducing the performance of the model.
Therefore, we add initial residual connections and identity mapping, following the work \cite{chen_2020}. 
%Since these mechanisms retain the unsmoothed feature as the initial representation of an entity, the entity representations at layers can be distinguished. 
Formally, the formula of residual connection and identity mapping can be described as:
%\begin{align}
%h_{i}^{(l+1)} = \sigma[((1-\beta_{l})\mathbf{I}_{n}+ \beta_{l}\mathbf{W}_{i}^{(l)}) \cdot ((1-\alpha_{l})\Tilde{\mathbf{P}}h_{i}^{(l)} + \alpha_{l}h_{i}^{(0)})] 
%\end{align}
\begin{align}
H^{(l+1)} =& \sigma\left[\left((1-\alpha_{l})\Tilde{P}H_{i}^{(l)} + \alpha_{l}H_{i}^{(0)}\right) \left((1-\beta_{l})I_{n}+ \beta_{l}W^{(l)}\right)\right] ,
\end{align}
where $\beta_{l} = \log{\frac{\lambda}{1+l}}$, $\Tilde{P} = (D + I_{n} )^{-1/2}(A + I_{n})(D + I_{n} )^{-1/2}$, and $\Tilde{P}$ is a 
graph convolution matrix, $\lambda$ and $\alpha_{l}$ are hyper-parameters. 
%$H^{(l)}$ is a $l$-th parameter matrix composed of $h_{*}^{(l)}$. \
%are concatenations of the learned linear transformed edge embeddings

%After computing the GAT layers, we concatenate the vectors obtained by attentive multi-layer. 
%By combining the initial layer with all the GNN layers, the whole graph structure can be leveraged and the model can remember the  initial embeddings \cite{wang_2019}. 
To compute the final representation of an $i$-th entity $e_{i}$, we first concatenate all the output representations of GNN layers. 
%By combining the initial layer with all the GNN layers, the whole graph structure can be leveraged and the model can remember the  initial embeddings \cite{wang_2019}.
As the local and global graph structures are important to represent entities, we aim to combine the initial feature with the output features of all GNN layers. 
Therefore, the representations could learn the local and global graph structures \cite{wang_2019, hong_2022_lagat}.
We then apply a linear function followed by an activation function to transform the entity vectors into final representations:

\begin{equation}
    e_{i} = \sigma\left(W \cdot \concat_{k=1}^K \left(h_{i}^{(k)}\right) + b\right) , 
\end{equation}
% \begin{equation}
%      = \sigma(\mathbf{W}e_{i} + \mathbf{b}) \ ,
% \end{equation}
where $K$ presents the number of GNN layers,
${b}$ is the bias of the linear function, and $W \in \mathbb{R}^{h \times K} \rightarrow {\mathbb{R}}^{d}$ is the weight matrix.% to transform the final embedding into a dimension $d$.
% \begin{equation}
%     \hat{y} (h,t) = \langle \phi(h), \phi(t)\rangle
% \end{equation}
% where $\hat{y}(h,t) \in [0,1]$ indicates similarity between embedded two node $u$ and $v$.

% Since entity embeddings could be used in various downstream tasks (e.g., disease diagnosis, disease classification, or even medicine recommendations),
% we 

% such as disease diagnosis, disease classification, or even medicine recommendations developed to solve health issues in the real world. 

%Finally, the final embeddings are continuously trained for downstream tasks to solve several medical applications, such as disease diagnosis, classification, or drug recommendations. To flexibly transfer downstream tasks, we define \textit{pre-training} and \textit{fine-tuning} phases \cite{han2021adaptive} that are described in detail in the following sections.

% \subsection {Learning Multiple-Relations in Knowledge Graph}
\subsection {Pre-training with Focusing on Multi-relational Structures}

%Knowledge graph embedding methods can be divided into translation-based, matrix factorization-based, and neural network-based models based on the embedding technique \cite{dai_2020_survey}. Embedding models have different characteristics that can be used to deal with different graphs. Hence, the aspects of the graph structure are considered to choose a suitable model. Since our MKG contains several types of relations containing one-to-one, one-to-many, and many-to-many, it is challenging to leverage these complicated relations. 
We now represent our pre-training task to learn multi-relational structures between different types of entities in KG.
In this study, we aim to preserve the triplet relations using a translation-based method.
For each triplet, entity embedding vectors are first transformed to a shared space through a projection matrix. 
Then, we use a triplet score function to calculate their relation score \cite{lin_2015}. 
%As a result, relation types of pair entities are learned independently, and the model can represent the complicated types of relations mentioned above. 
To preserve all the entity relations, we aim to maximize all the positive triplets coming from KGs and minimize all the negative triplets that are not coming from KGs \cite{bordes_2013,lin_2015}. 
%This loss function maximizes the relation proximity of a correct triplet while minimizing the proximity of an incorrect triplet. 
Formally, the scoring function is defined as:
\begin{equation} 
 f_{score} = \hat{y}(h,r,t) = W_rh + r -W_rt \ ,
\end{equation}
where $h$, $t$, and $r$ are the output representations of head, tail, and relation, and $W_r$ is a projection matrix to map entities and relations onto the $r$ space. 
A triplet loss function is computed to compare the positive and negative triple pairs defined as:
\begin{align} 
 \mathcal{L}_\mathcal{P}(\mathcal{T}) = \sum_{\forall(h,r,t) \in \mathcal{T}}&-{\ln\sigma\left(\hat{y}(h,r, \Bar{t} ) - \hat{y}(h,r,t)\right)} \notag+ \lambda \vert \vert  \Theta \vert \vert _2 ^2 , \\
\end{align}
where $\mathcal{T}$ denotes the set of triplets, $(h,r, \Bar{t})$ denote the negative triplet, $\Theta$ refers to $L_2$ regularization parameter.  

\subsection{Fine-tuning for Animal Disease Diagnostics}

%Medical applications (such as disease diagnosis, medicine recommendation, and disease classification) can be developed to solve real-world tasks by continuously training the pre-trained embeddings \cite{baykal_2020_transfer, lim_2020_classification}. 
We now apply learned LiteralKG and representations to new downstream task, such as predicting animal diseases.
%We train binary classification to predict the missing information in a new medical record. 
%For disease diagnosis, the model predicts correct diseases for a new medical record, known as a newly injected sub-graph. 
%When a new patient comes to the hospital, doctors will input a new medical record that does not have the disease, treatment, and drug information. 
%The model then predicts the relations between the medical record entity and the missing entities, such as diseases or drugs. 
We first compute a coefficient score to measure the relationship between each head and tail pair. 
The equation for calculating the coefficient score can be formulated as follows:
\begin{equation}
    \hat{y}_{h,t} = \hat{y} (h,t) = \langle \phi(h), \phi(t)\rangle \ = FC(W_rh || W_rt) ,
\end{equation}
where $h$ and $t$ denote the learned representations of head and tail, respectively, $||$ is the concatenation operator, $W_r$ is the transformation matrix corresponding to the relation between $h$ and $t$, and $FC$ denotes the fully connected layers to get the prediction output.

For disease diagnosis, the model classifies the coefficient score between two entities into binary digits. With pre-trained embeddings, two observed entity vectors are first projected into the relation space. Their embedding vectors are then concatenated and transformed into a one-dimensional class by MLP layers to form the coefficient score. It is used to identify the relationship probability among the observed entities. Additionally, a binary cross entropy loss function is utilized for training classification. The loss function is described as:
\begin{align} 
 \mathcal{L}_\mathcal{F}(\mathcal{M}) = & -\sum_{
\forall(M_k, D_i) \in \mathcal{M}} \left[ y_{M_k, D_i}\log\left(\hat{y}_{M_k, D_i}\right) +  ( 1 - y_{M_k, D_i})\log\left((1-\hat{y}_{M_k, D_i})\right) \right] \ ,
\end{align}
where $M_k$ and $D_i$ are the medical record and disease entity, $y_{M_k, D_i}$ and $\hat{y}_{M_k, D_i}$ are the actual and predicted outputs, and $\mathcal{M}$ is the collection of all the training medical records containing positive and negative disease information.

% Since a medical record contains the patient's information and correct diseases, no negative samples exist in the training data. Therefore, a strategy is proposed to sample negative diseases for a record (a subgraph). In particular, there are positive diseases in a subgraph. Corresponding to each positive disease, the negative samples can be sampled by randomly selecting the others in the disease set, ignoring that positive entity. There is a rate of choosing negative entities. In this experiment, 3 negative samples are selected for each positive sample to reduce the prediction of wrong diseases for a patient.

% From the learned representations, other tasks can be solved by replacing the loss function. 
% For example, in drug recommendation, the BPR \cite{rendle_2012_bpr} loss can be exploited to maximize the proximity score between the medical record and correct drug entities. 
% Then, the proximity score of pair entities can be ranked to find the most efficient drugs. 
% Therefore, an early advantage of the pre-trained model can be aware of reducing time consumption when utilizing a large dataset for many applications.

%%%%%%%%%%%%%%%%%%%%%%%%%%%%%%%%%%%%%%%%%%
\section{Experiments}
\label{section_5}

In this section, we provide extensive experimental results to validate the performance of our model versus baselines.
In addition, we conduct ablation studies to investigate the contribution of the combination of different types of relations as well as residual connection and identity mapping to the overall performance.
%the proposed method, including a dataset description, experimental settings, and baselines. 
% we aim to evaluate the model in different settings.
%Therefore, we determine the most effective agg using several experimental strategies, such as evaluating different types of literals, applying residual connection and identity mapping, and the number of GAT layers. 
% From these results, we aim to determine the following questions:

% - Which model will fit the current MKG structure?

% - Is extracting rich data from literals highly effective in an MKG?

\subsection{Experimental Settings}

%In this section, the settings and procedures of the experiments are described. The below subsections are the dataset description, evaluation metrics, baseline models, and detailed settings.

\subsubsection{Dataset Description}

% We collected the data from 31 animal hospitals to construct our own training and testing datasets. In particular, the data includes 481,453 electric medical records that have been cleaned and are suitable for applying the AI model's duties. Based on this data, we constructed our Knowledge Graph containing 3,176,751 entities and 16 relations. Specifically, there are 2 phases in our model, including pre-training and fine-tuning. Based on these above conditions, We contributed all data to satisfy these phases by 80\% for pre-training and 20\% for the other. Additionally, the training and testing data were divided with a ratio of 4 and 1, respectively. As a result, we distributed 304,000 records for training in the pre-training phase and 76,000 for its testing. For the remaining data, we used 81,163 records for training and 20,290 records for testing the fine-tuning phase. In the fine-tuning phase, We built the testing dataset by extracting several medical records that are missing disease information, based on the concept of anticipating broken links in medical records to perform disease diagnosis. 

As mentioned earlier, we collected 85,965 EMRs from 31 animal hospitals and transformed to a knowledge graph. 
%In particular, 85,965 electronic medical records have been cleaned and are suitable for applying the AI model's duties. Based on this data, an MKG containing 595,172 entities and 16 relations is constructed to solve the link prediction task. 
After transforming, our MKG contains a total of 595,172 entities and 16 relation types. 
%Specifically, there are 2 phases in our model, including pre-training and fine-tuning. 
%Based on the above conditions, the constructed dataset is distributed into 80\% for the former and 20\% for the latter.
Note that we sampled three negative triplets for each positive triplet in the experiments.
We first pre-trained our model in a self-supervised manner without using any label information.
Then, we fine-tuned the LiteralKG model to learn the knowledge for link prediction tasks.
For the fine-tuning task, we conducted the experiment by randomly sampling training, validation, and testing sets of size 60\%, 20\%, and 20\%, respectively.

%Additionally, the training and testing data were divided with a ratio of 4 and 1, respectively. 
%As a result, we distributed 55,013 records for training and 13,829 for testing of pre-training. 
%For the remaining data, we used 13,756 records for training and 3,367 records for testing of fine-tuning. 
%In the fine-tuning phase, We built the testing dataset by extracting several medical records that are missing disease information, based on the concept of anticipating broken links in medical records to perform disease diagnosis. 	

\subsubsection{Evaluation Metrics}

%This study aims to anticipate broken links in a new medical record. For each new record, we identify the disease by classifying the relation of the center entity $M_k$ with the list of diseases. An animal has a disease when there is a relation between its medical record and this disease. 
Since our task is a binary classification problem, we utilized several evaluation metrics, including accuracy ($Acc$), precision ($P$), recall ($R$), and $F_1$. 
The evaluation metrics are defined as follows:
\begin{align}
    Acc = \frac{|D^+_{true} \cup D^-_{true}|}{|D^+ \cup D^-|} \ , \ 
    % Precision = \frac{|D \cap P|}{|P|} \ ,\\
    % Recall = \frac{|D \cap P|}{|D|} \ , \ 
    P = \frac{|D^+_{true}|}{|D^+|} \ , 
    R = \frac{|D^+_{true}|}{|D^+_{true} \cup D^-_{false}|} \ , \      
    F_1 = 2 \frac{(P R)}{(P + R)} \ ,
\end{align}
where the $D^+_{true}$ and $D^-_{true}$ denote the correct predictions for positive and negative diseases, respectively, and $D^{-}_{false}$ is the incorrect predictions for negative diseases.  
These criteria are intended to assess how well our model performs compared to baselines. 
%Additionally, we assess how the model's performance has changed due to changing the model characteristics and features that we have incorporated, such as pre-training influence, residual connection efficacy, exploiting literals, and the influence of GNN layers.

\subsubsection{Baselines}

We compare our model to relevant translation-based methods and GNN models, which have gained remarkable success in KG representation learning.
Translation-based models learn representations by mapping the entities and their relations into latent space through translations from head-to-tail entities.
\begin{itemize}
  \item \textbf{TransE} \cite{bordes_2013}. The model uses a simple scoring function to compare the relationships between the pairs of entities. 
  Note that all types of entities and relations are represented in the same space. 
  % Given $y_{h}$ and $y_{t}$ denotes the embedding of head and tail, r is the relation vector, we construct a scoring function of TransE as follows:
  %       \begin{equation}
  %           f^{TransE}_r(y_{h}, y_{t}) =\parallel y_h + r - y_t \parallel
  %       \end{equation}
  \item \textbf{TransR} \cite{lin_2015}. TransR uses projection matrices to map various types of entities into different relation spaces and then construct translations between entities.

   \item \textbf{SMR} \cite{gong_2021}. The idea of SMR is to use TransR to transform entities into latent space linearly.
   Meanwhile, SMR uses the LINE \cite{tang_2015_line} model to capture the similarity between entities, which successfully addresses homogeneous graphs.

   %represent the relations in MKG and to leverage the bipartite graph structure.
    % \item \textbf{Disease Prediction via GNN} \cite{sun_2021} uses GAT and GIN to encrypt the graph structure for disease prediction while retaining both the medical concept graph(or MKG) and patients' diagnostic information in the patient record graph.
    % \item \textbf{PoLo} \cite{liu_2021} embeds the KG structure using logical meta-paths generated based on the policy-guided walks. As a result, this method can keep the logical structure while learning the MKG representation.
    \end{itemize}

Furthermore, we also compare our model with recent GNNs, which could learn high-order sub-structures and semantic relations in KGs.
There are three GNN baselines, including KGNN, KGNMDA, and LaGAT model, as:

    \begin{itemize}
    
    \item \textbf{KGNN} \cite{lin_2020_kgnn} learn KG structures through a local receptive to aggregate neighbour features and topological information.
    KGNN could also capture the high-order structures surrounding target entities and semantic relations to learn global structural information.
    % \item \textbf{SumGNN} \cite{yu_2021_sumgnn} 
    % \item \textbf{DDKG} \cite{su_2022_attention} constructs a biomedical knowledge graph to predict the drug-drug interaction based on graph neural network aggregation and attention mechanism.
    \item \textbf{KGNMDA} \cite{jiang2022kgnmda} represent the relations between microbes and diseases based on the Gaussian kernel and then learn the similarity between them in an uncertain manner.
    They then use a linear transformation to predict the scores across microbe-disease relations.
    
    \item \textbf{LaGAT} \cite{hong_2022_lagat} extends KGNN by using an attentive mechanism, which could learn different weighted messages contributed from different entities. 
    Furthermore, the outputs of different attentive embedding propagation layers are concatenated and then contribute to the final representations.
    
  %various types of relations in a KG by projecting the entities represented in their own space onto a separate space known as the relation space by using a projection matrix.
        % \begin{equation}
        %     f^{TransR}_r(y_{h}, y_{t}) = 	\parallel y_h\textbf{M}_r + r - y_t\textbf{M}_r \parallel
        % \end{equation}
 
  % \item \textbf{KG-Predict} \cite{gao_2022} captures the variety of heterogeneous interactions retained in the entity and relation embedding by utilizing the features of both the entities and the sets of interactions.
\end{itemize}

\subsubsection{Implementation Details}

Our model is implemented based on the Pytorch library.
Adam optimizer \cite{kingma_2014_adam} was applied to the pre-training and fine-tuning phases. We applied Leaky ReLU as an activation function in the aggregation layers. Additionally, the dimensions of entity and relation vectors were set to 300. We initialized the learning rate at 0.0001. Referred to current GNN hyper-parameter settings \cite{kipf_2016, hamilton_2017, xu_2018, wang_2019}, the tuning hyper-parameters are: 
(1) the hidden dimension of the GNN layer $\in \{16, 32\}$, (2) the batch size for the pre-training and fine-tuning phases $\in \{1024, 2048\}$, 
(3) the number of GNN layer $\in \{2, 4, 8, 16\}$, 
and (4) the dropout ratio $\in \{0.1, 0.5\}$ after each layer. The number of layers in various GNN models was tuned to evaluate the performance of these models and the effectiveness of applying residual connection in MKG.
%The source code has been published at \url{https://github.com/NSLab-CUK/LiteralKG}.
For fair comparisons with the baselines, the hyper-parameters in all baselines were tuned in the same range, including the learning rate $\in \{0.001, 0.0001\}$, hidden dimension $\in \{16, 32, 64, 128\}$ and the number of layers $\in \{1, 2, 4\}$. 
%For KGNMDA, which generates Gaussian kernel similarity features, there is one more hyper-parameter, Gaussian feature dimension $\in \{32, 64\}$. Moreover, the highest value of $F_1$ measure was chosen to identify an early stopping point for all the baselines and the proposed model.

\subsection{Performance Analysis}

%Several KG models are available for the link prediction task in general and disease diagnosis in MKG. However, choosing a model appropriate for the graph's structure requires careful consideration of many factors. We conducted research, contrasted various approaches, and analyzed the performance of some models to examine and evaluate the usefulness of KG embedding methods on MKG.

\begin{table}[t]
\centering
\caption{The performance of LiteralKG and baselines in terms of accuracy, recall, precision and $F_1$. 
The top two are highlighted by \textcolor{red}{first} and \underline{second}.}
\vspace{0.1in}
  \begin{tabular}{l  cccc}
  \toprule
\textbf{Model} & \textbf{Accuracy}	& \textbf{Recall} & \textbf{Precision} & \textbf{$F_1$}
\\\hline
TransE  \cite{bordes_2013}	    & 0.5937    & 0.5617  & 0.6001    & 0.5802 \\
TransR  \cite{lin_2015}		    & 0.5732    & 0.5915  & 0.5706    & 0.5808 \\
SMR	\cite{gong_2021}	    & 0.5872    & 0.5669  & 0.5909    & 0.5787 \\
KGNN	\cite{lin_2020_kgnn}	    & 0.7890    & 0.7021  & 0.8499   & 0.7689 \\
KGNMDA \cite{jiang2022kgnmda}		    & 0.8130    & 0.7111  & \underline{0.8932}   & 0.7918 \\
LaGAT	\cite{hong_2022_lagat}	    & \underline{0.8545}    & \underline{0.7988}  & \textcolor{red}{0.8988}   & \underline{0.8459} \\
LiteralKG		   & \textcolor{red}{0.8616}	& \textcolor{red}{0.9357} & 0.8150 & \textcolor{red}{0.8712} \\\bottomrule
\end{tabular}

%The first and second performances are marked by \textbf{bold} and \underline{underline}, respectively.}
\label{tab:3}

\end{table}

Table~\ref{tab:3} shows the performance of our model and baselines in terms of accuracy, recall, precision, and $F_1$.
We have the following observations:
\textbf{(1)} LiteralKG with pre-training outperformed baselines in most measurements.
%There is a significant improvement when leveraging the attribute information, combining the triplet score of KG with GAT, and concatenating all GNN layers in our model compared to the other baselines. 
Specifically, our proposed model reached the best value at $0.9357$ regarding Recall measurement. 
We argue that our model could learn the literal information well to maximize the relations between entities. 
%It benefits applications that concentrate on predicting the proximity among the entities in KG.
%Regarding disease diagnosis, the proposed model focuses on accurately predicting patients' disease rather than predicting animals who are unlikely to have that disease. 
 %Therefore, GNN models, which aim to maximize precision performance following the negative rate, show a higher precision. On the other hand, the negative samples are not recognized and can be incorrect. 
%As a result, with the same dataset for all the evaluated models, the proposed model can leverage the semantic similarity of textual features to recognize these incorrect negative samples, which hinders the precision score compared to others.
% Due to uncertain negative samples, our model shows lower recall than the other GNN models, aiming to maximize the performance based on the datasets. 
\textbf{(2)} Translation-based models, i.e., TransE, TransR, and SMR, showed low performance in predicting disease.
We argue that these models overlooked literal information and eventually could not capture the complex relations since they only learn representations through simple linear transformations. 
%With a large-scale graph and star structure, leveraging neighbour information is important to predict unseen entities. 
\textbf{(3)} LaGAT showed competitive performance compared to our model.
We argue that as LaGAT could learn graph structure through the attention mechanism, the model thus could learn attentive weights contributed from different neighbourhood entities in KGs.
%However, without learning literal information, the overall performance could then not compara
%the graph structure using GNN and attention mechanism to distinguish noise messages, this model shows higher performance than KGNN and KGNMDA and archives second performance on the accuracy, recall, and $F_1$ measure. However, our model outperforms all the baselines because of the valuable information in attributes and the power of the triplet loss function in the pre-training phase. 

\subsection{Ablation Studies}

%In this work, we employ a gating mechanism \cite{cho_2014_learning} to capitalize on the extensive range of attributes. Therefore, we evaluate the effectiveness of using the literal in disease diagnosis.

\begin{table}
\centering
\caption{The results for evaluating the efficiency of leveraging literals. The first column contains the combining literal strategies. There are four cases, including without (w/o) literal, without textual, without numerical, and the combination of numerical and textual features.
The top two are highlighted by \textcolor{red}{first} and \underline{second}.}
\vspace{0.1in}
\label{tab:4}

  \begin{tabular}{l  cccc}
\toprule
\textbf{Combining literals} & \textbf{Accuracy}	& \textbf{Recall} & \textbf{Precision} & $\textbf{F}_1$\\
\midrule
w/o \ literal		    & 0.8223  & 0.8664   & 0.7962     & 0.8298 \\
w/o \ textual	    & 0.8424    & 0.8544     & \textcolor{red}{0.8344} & 0.8443 \\
w/o \ numerical		    & \underline{0.8529}    & \underline{0.9116}     & \underline{0.8158} & \underline{0.8610} \\
numerical $+$ textual		    & \textcolor{red}{0.8616}	& \textcolor{red}{0.9357} & 0.8150 & \textcolor{red}{0.8712}  \\
\bottomrule
\end{tabular}
\end{table}

\subsubsection{On the Importance of Learning Literal Information}

Table~\ref{tab:4} shows how the literal information contributes to the overall performance of our model.
We have the following observations:
\textbf{(1)} LiteralKG showed the best performance with the combination of numerical and textual information in most measurements. 
We argue that both types of literal information could benefit our model to explore similar entities in the KG and thus contribute to the overall performance. 
%On the one hand, removing the textual feature decreases the model performance from 0.8616 to 0.8424 in the accuracy measure and 0.8712 to 0.8443 in the $F_1$ measure. 
\textbf{(2)} We find out that without numerical information, the performance of our model increased slightly, from 0.8529 to 0.8616 and from 0.8610 to 0.8712 in terms of accuracy and $F_1$, respectively. 
It indicates that the textual information contributes considerably to the overall performance compared to numerical attributes.
Thanks to the textual features, LiteralKG performs well even without numerical features.
%Additionally, recall receives the lowest value of 0.8544 when ignoring the textual features, showing the efficiency of incorporating textual features.
%Otherwise, the prediction is improved when adding the literal information. 
%It can be explained that there are more entities containing textual attributes than entities containing numerical attributes in our MKG. $F_1$ measure shows the lowest performance when ignoring all the attribute features. 
\textbf{(3)} When combining different types of attributes, the recall measurement increased from 0.8664 to 0.9357, which shows a good point for maximizing the ability to diagnose patients with a correct disease. 
It indicates that the gating mechanism transformed the attributes and entity embedding vectors into the shared space and eventually contributed to the overall performance. 
% Comparing Precision and Recall results, the model leveraging textual attributes archives better performance from 0.8544 to 0.9357 in Precision and decreases F1 from 0.8344 to 0.8150. 
% The reason can be observed that the textual features are embedded using Fasttext, which leverages semantic similarity among texts. Therefore, it improves the model performance in predicting similar entities.

\begin{table}[t]
\centering
\caption{
Link prediction performance on the Residual Connection and Identity Mapping (R.C\& I.M) in GNN aggregators. $\times$ and $\bigcirc$ denote a setting ignoring and adopting residual connection, respectively. 
%The results compare the performance of 4 GNN aggregators, including GCN, GraphSAGE, Bi-Interaction, and GIN.
The top two are highlighted by \textcolor{red}{first} and \underline{second}.}
\vspace{0.1in}
\label{tab:5}
\begin{tabular}{l  ccccc}
\toprule

\textbf{R.C\& I.M} & \textbf{Aggregator} & \textbf{Accuracy}	& \textbf{Recall} & \textbf{Precision} & $\textbf{F}_1$\\
\midrule
\multirow{4}{*}{$\times$}
& GCN		     & 0.8532 & \textcolor{red}{0.9419} & 0.8000 & \underline{0.8652} \\
& GraphSAGE	     & \textcolor{red}{0.8580} & \underline{0.9257} & \underline{0.8154} & \textcolor{red}{0.8670} \\
& Bi-Interaction & \underline{0.8545} & 0.9123 & \textcolor{red}{0.8178} & 0.8625 \\
& GIN		     & 0.8387 & 0.8621 & 0.8236 & 0.8424 \\

\bottomrule
\multirow{4}{*}{$\bigcirc$}
& GCN		    & \underline{0.8527} & 0.9170 & \underline{0.8125} & \underline{0.8616} \\
& GraphSAGE		   & 0.8300 & 0.8882 & 0.7956 & 0.8394 \\
& Bi-Interaction	& \textcolor{red}{0.8616} & \underline{0.9357} & \textcolor{red}{0.8150} & \textcolor{red}{0.8712} \\
& GIN	   & 0.8490    & \textcolor{red}{0.9372}     & 0.7967 & 0.8612 \\
\bottomrule
\end{tabular}
\end{table}

% \begin{tabularx}{0.48\textwidth}{p{0.11\textwidth}p{0.06\textwidth}p{0.06\textwidth}p{0.06\textwidth}p{0.08\textwidth}}

\begin{table}[t]
\centering
\caption{
The performance of LiteralKG on the role of the pre-training task. $\times$ and $\bigcirc$ denote a setting without a pre-trained model and utilizing the pre-trained model, respectively.
The top two are highlighted by \textcolor{red}{first} and \underline{second}.
}
\label{tab:6}
\vspace{0.1in}

\begin{tabular}{l  ccccc}

\toprule
\textbf{Pre-training} & \textbf{Aggregator} & \textbf{Accuracy}	& \textbf{Recall} & \textbf{Precision} & $\textbf{F}_1$\\
\midrule
\multirow{4}{*}{$\times$}
& GCN		     &\textcolor{red}{0.8183}     & 0.7371             & \textcolor{red}{0.8801}     &\textcolor{red}{ 0.8022} \\
& GraphSAGE	     & \underline{0.8109} & \underline{0.7429} & 0.8597              & 0.7971 \\
& Bi-Interaction & 0.7976             & 0.7083             & \underline{0.8622}  & 0.7777 \\
& GIN		     & 0.8013             & \textcolor{red}{0.8053}    & 0.7989              & \underline{0.8021} \\
\bottomrule
\multirow{4}{*}{$\bigcirc$}
& GCN		    & \underline{0.8527} & 0.9170 & \underline{0.8125} & \underline{0.8616} \\
& GraphSAGE		   & 0.8300 & 0.8882 & 0.7956 & 0.8394 \\
& Bi-Interaction	& \textcolor{red}{0.8616} & \underline{0.9357} & \textcolor{red}{0.8150} & \textcolor{red}{0.8712} \\
& GIN	   & 0.8490    & \textcolor{red}{0.9372}     & 0.7967 & 0.8612 \\
\bottomrule
\end{tabular}

\end{table}

\subsubsection{On the Importance of Residual Connection and Identity Mapping}

We further conducted experiments to evaluate the effectiveness of residual connection and identity mapping shown in Table~\ref{tab:5}. 
We have the following observations:
\textbf{(1)} The overall performance increased slightly by applying the residual connection and identity mapping. 
It indicates that even though KGs have complex relations and heterophily properties, residual connections and identity mapping could act as auxiliary modules to contribute to the overall performance of LiteralKG.
In other words, the two modules could help LiteralKG prevent over-smoothing problem and improve the performance of our model. 
\textbf{(2)} In comparison with GraphSAGE, the performance of our model with residual connection has been reduced from 0.8580 to 0.8300 and from 0.8670 to 0.8394 in terms of accuracy and $F_1$, respectively. 
It indicates that the residual connection does not contribute much to our model performance as GraphSAGE sampled the neighbourhoods based on random walks.
%Since we apply residual connection and identity mapping to overcome the vanishing gradient problem of a deep GNN, 
%Ignoring the residual connection, we compare this model performance with the incorporating residual connection version.
%while, on the one hand, the residual connection can not show considerable improvement in a small number of layers.  
%On the other hand, in the final step, all the features of GNN layers are combined with the initial feature, which can be seen as one way to overcome the over-smoothing problem when extracting the graph structure in %deep GNN layers. These results solve the concern in RQ 3.
%using the residual connection in different GNN aggregators can show different effects. 
%Unlike GCN, Bi-Interaction, and GIN aggregators, which combine neighbor features with the node features of the previous layer, the concatenation aggregator on GraphSAGE distinguishes these features, which shows a lower performance when adding the initial node feature. In contrast, Bi-Interaction contains two linear transformation weight matrices, which raise a positive effect in combining with residual connection settings. 
\textbf{(3)} For the GIN aggregator, using residual connection can improve the model performance from 0.8174 to 0.8378 in terms of $F_1$ measure. 
Note that GIN aggregators aim to map different sub-structures into different representations, leading to the power of 1d-WL isomorphism testing.
This implies that residual connections could contribute to the model performance as GIN aggregators could ignore the original features of entities.

%Without residual connection, the GraphSAGE aggregator archives the best performance on $F_1$ and accuracy, demonstrating more power when capturing the graph structure in this setting. 
%Otherwise, since the sum aggregation in Bi-Interaction and GCN combine with residual connection and identity mapping in GNN layer, these aggregators outperform others on accuracy, $F_1$, and precision. 

\begin{center}
\begin{figure}
\centering
\resizebox{14cm}{5cm}{
\begin{tikzpicture}
\tikzstyle{every node}=[font=\normalsize]
\begin{axis}[
        xlabel=$Number\ of\ layers$,
        ylabel=$Accuracy\ (\%)$,
        ymin=60, ymax=100,
        ytick={60,70,80,90,100},
        symbolic x coords={1,2,4,8},
        xtick=data,
        legend pos=south west,
        legend cell align={left}
        ]
    \addplot[mark=*,cyan] coordinates {
        (1,85.27)
        (2,83.29)
        (4,81.17)
        (8,72.89)
    };
    \addlegendentry{ \normalsize GCN}

    \addplot[color=red,mark=x]
        coordinates {
            (1,82.57)
            (2,83.00)
            (4,77.75)
            (8,75.77)
        };
        \addlegendentry{ \normalsize GraphSAGE}

    \addplot[color=green,mark=+]
        coordinates {
            (1,86.16)
            (2,80.28)
            (4,73.93)
            (8,69.10)
        };
        \addlegendentry{ \normalsize Bi-Interaction}

    \addplot[color=orange,mark=square*]
        coordinates {
            (1,84.47)
            (2,84.90)
            (4,83.97)
            (8,82.57)
        };
        \addlegendentry{ \normalsize GIN}
    \end{axis}
\end{tikzpicture}
\begin{tikzpicture}
   \tikzstyle{every node}=[font=\normalsize]
    \begin{axis}[
        xlabel=$Number\ of\ layers$,
        ylabel=$F_1$,
        ymin=0.6, ymax=1,
        symbolic x coords={1,2,4,8},
        xtick=data,
        ytick={0.6,0.7,0.8,0.9,1},
        legend pos=south west,
        legend cell align={left}
        ]
    \addplot[mark=*,cyan] plot coordinates {
        (1,0.8616)
        (2,0.8474)
        (4,0.8212)
        (8,0.7494)
    };
    \addlegendentry{ \normalsize GCN}

    \addplot[color=red,mark=x]
        plot coordinates {
            (1,0.8375)
            (2,0.8394)
            (4,0.7801)
            (8,0.7677)
        };
        \addlegendentry{ \normalsize GraphSAGE}

    \addplot[color=green,mark=+]
        plot coordinates {
            (1,0.8712)
            (2,0.8136)
            (4,0.7455)
            (8,0.7100)
        };
        \addlegendentry{ \normalsize Bi-Interaction}

    \addplot[color=orange,mark=square*]
        plot coordinates {
            (1,0.8537)
            (2,0.8612)
            (4,0.8508)
            (8,0.8379)
        };
        \addlegendentry{ \normalsize GIN}
    \end{axis}
\end{tikzpicture}
}
\centering
\resizebox{14cm}{5cm}{

\begin{tikzpicture}
    \tikzstyle{every node}=[font=\normalsize]
    \begin{axis}[
        xlabel=$Number\ of\ layers$,
        ylabel=$Recall\ (\%)$,
        ymin=60, ymax=100,
        symbolic x coords={1,2,4,8},
        xtick=data,
        ytick={60,70,80,90,100},
        legend pos=south west,
        legend cell align={left}
        ]
    \addplot[mark=*,cyan] plot coordinates {
        (1,86.16)
        (2,84.74)
        (4,82.12)
        (8,74.94)
    };
    \addlegendentry{ \normalsize GCN}

    \addplot[color=red,mark=x]
        plot coordinates {
            (1,89.84)
            (2,88.82)
            (4,78.96)
            (8,80.08)
        };
    \addlegendentry{ \normalsize GraphSAGE}

    \addplot[color=green,mark=+]
        plot coordinates {
            (1,93.57)
            (2,86.06)
            (4,76.37)
            (8,75.65)
        };
    \addlegendentry{ \normalsize Bi-Interaction}

    \addplot[color=orange,mark=square*]
        plot coordinates {
            (1,90.66)
            (2,93.72)
            (4,91.38)
            (8,90.14)
        };
    \addlegendentry{GIN}
    \end{axis}
\end{tikzpicture}
\begin{tikzpicture}
    \tikzstyle{every node}=[font=\normalsize]
    \begin{axis}[
        xlabel=$Number\ of\ layers$,
        ylabel=$Precision\ (\%)$,
        ymin=60, ymax=100,
        symbolic x coords={1,2,4,8},
        xtick=data,
        ytick={60,70,80,90,100},
        legend pos=south west,
        legend cell align={left}
        ]
    \addplot[mark=*,cyan] 
        plot coordinates {
            (1,81.25)
            (2,77.95)
            (4,78.17)
            (8,69.68)
        };
        \addlegendentry{\normalsize GCN}

    \addplot[color=red,mark=x]
        plot coordinates {
            (1,78.43)
            (2,79.56)
            (4,77.09)
            (8,73.73)
        };
        \addlegendentry{\normalsize  GraphSAGE}

    \addplot[color=green,mark=+]
        plot coordinates {
            (1,81.50)
            (2,77.15)
            (4,72.82)
            (8,66.88)
        };
        \addlegendentry{\normalsize  Bi-Interaction}

    \addplot[color=orange,mark=square*]
        plot coordinates {
            (1,80.67)
            (2,79.67)
            (4,79.59)
            (8,78.28)
        };
        \addlegendentry{\normalsize  GIN}
    \end{axis}
\end{tikzpicture}
}
\centering
\caption{ Performance on link prediction over attentive embedding propagation layers on four types of aggregators, including GCN, GraphSAGE, Bi-Interaction, and GIN aggregators.}
\label{fig:4}
\end{figure}
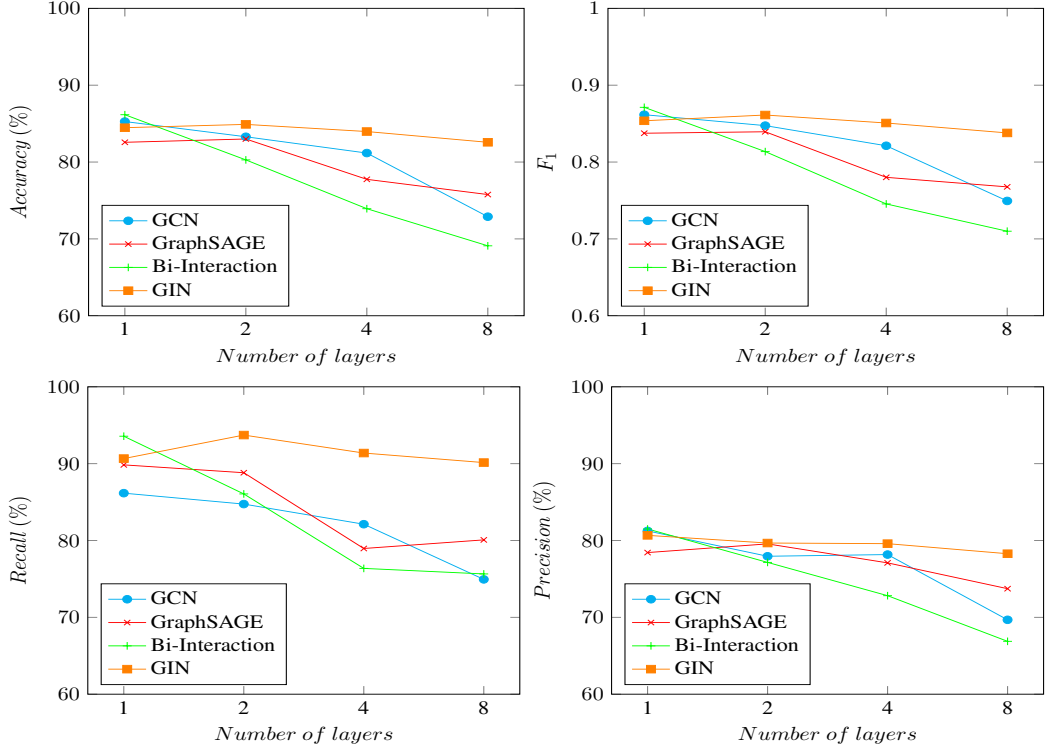
\end{center}

\subsubsection{On the Pre-training Phase}

%There are pre-training and fine-tuning phases in this architecture. In this section, we investigate the effects of pre-training on disease diagnosis performance. To evaluate the influence of the pre-training phase, we ignored it and checked the performance of directly training the model with the fine-tuning loss function. 
Table \ref{tab:6} shows the link prediction performance on the pre-training phase evaluation. 
We discovered that our pre-trained model could show comparable performance in most of the aggregators. 
In particular, LiteralKG reaches the highest values in the Bi-Interaction aggregator at 0.8616, 0.8150, and 0.8712 in terms of accuracy, precision, and $F_1$, respectively. 
%Moreover, in Bi-Interaction, the model's accuracy grows from 0.7976 to 0.8616 when combining the pre-training phase. 
%Looking at the second performance in adapting the pre-training phase, $F_1$ measure and accuracy in GCN  increase from 0.8022 to 0.8616 and 0.8183 to 0.8527, respectively. 
We argue that the triplet scoring function in the pre-training phase first learns to maximize relation proximity among entities in correct triplets. 
Therefore, it improves the performance of the fine-tuning model in link prediction, in which the link is also considered by representation proximity. 
In GIN, applying the pre-trained model improves performance on $F_1$ and accuracy from 0.8021 to 0.8612 and 0.8013 to 0.8490, respectively. 
%Besides this improvement, when adopting a pre-training phase, the learned representations could promptly utilized to construct other applications, such as disease classification or drug recommendation, without retraining the whole graph data. 
These results, therefore, explain the efficiency of the pre-trained model, which satisfies the RQ2.

\subsection{Sensitivity Analysis}

Figure~\ref{fig:4} shows the performance of our model by applying different range of GNN layers. 
We have the following observations:
\textbf{(1)} This result illustrates that LiteralKG with a $1$-layer or $2$-layer can efficiently learn the KG structure in most of the aggregators. 
%In a star network, medical record information can be leveraged in a small number of GNN layers \cite{chen_2020}. 
For the Bi-Interaction aggregator, LiteralKG achieves the best performance when using only one layer, which captures 1-hop neighbourhood entities. 
We argue that as Bi-Interaction could capture the graph structure through sum aggregators and element-
wise product, the aggregators could suffer the over-smoothing problem when staking more layers.
In other words, Bi-Interaction aggregators could not distinguish sub-structures as the aggregators learned the similarity between target nodes and neighbours at the first layers.
%Otherwise, GraphSAGE and GIN archive first and second performances in 2-layer and 1-layer GNN, respectively. 
%The main difference between these types of aggregators can cause these results. 
%GCN and Bi-Interaction mix the neighbour features with the entity feature, which can leverage the graph structure in a small number of GNN layers in a high-degree entities graph. 
%Additionally, the model performance drops slightly when increasing the number of GIN layers. 
%It demonstrates the ability of GIN to deal with the over-smoothing problem due to the dense connection of this aggregator.
\textbf{(2)} It is worth noting that GIN aggregators show the best performance when staking more GNN layers compared to other aggregators.
We argue that as the power of GIN achieves nearly 1d-WL isomorphism testing, the model then could handle the over-smoothing problem even staking more GNN layers.
%Otherwise, in Bi-Interaction aggregator, increasing the number of layers reduces the model performance faster than the others.
In other words, GIN aggregators could distinguish different sub-structures even staking more layers.
%The reason can come from the results of the element-wise product, which multiplies the features of all layers leading to an exponential growth in the value of hidden features at each layer so that it hinders the sum aggregator and faces the over-smoothing problem.

%\newpage

%%%%%%%%%%%%%%%%%%%%%%%%%%%%%%%%%%%%%%%%%%

%%%%%%%%%%%%%%%%%%%%%%%%%%%%%%%%%%%%%%%%%%
\section{Conclusion and Future Work}
\label{section_6}

In this study, we propose a knowledge graph embedding model, LiteralKG, which could learn different types of entity attributes to diagnose companion animal disease.
By doing so, we first constructed an MKG from various EMRs collected from 31 animal hospitals.
Then, LiteralKG fuses different types of literals into unified representations through a gating network.
We then use the attention mechanism with the initial feature to learn the coefficients across triplets
 to capture local and global graph structures. 
The experiment results show that our model outperforms the shallow KG embedding and GNN-based models due to the improvements from leveraging the literal features and the efficiency of the pre-trained phase.
Furthermore, we present a pre-training task that could learn graph structure and its properties without using any label information to generate learned representations.
The pre-trained model with representations could then be used for prediction tasks.
Besides, the negative samples from our data are sampled randomly so that it can affect the overall performance of LiteralKG. 
In future work, we plan to build useful sampling strategies to effectively build positive and negative samples, such as graph clustering and sub-graph sampling methods.

%In future work, we aim to build additional graph embedding models to handle several types of graphs, such as control flow or molecular graphs.
%Furthermore, we plan to integrate state-of-the-art graph transformer models into Connector.

\section*{Acknowledgments}
This work was supported in part by the National Research Foundation of Korea (NRF) grant funded by the Korea government (MSIT) (No. 2022R1F1A1065516 and No. 2022K1A3A1A79089461), in part by the Research Fund, 2023 of The Catholic
University of Korea (M-2023-B0002-00088), and in part by the Research Fund of IntoCNS (M-2022-D0827-00001).

%Bibliography
\bibliographystyle{unsrt}  
\bibliography{references}

\begin{thebibliography}{10}

\bibitem{zhang_2019_vettag}
Yuhui Zhang, Allen Nie, Ashley Zehnder, Rodney~L{\'{o}}pez Page, and James Zou.
\newblock Vettag: improving automated veterinary diagnosis coding via
  large-scale language modeling.
\newblock {\em NPJ digital medicine}, 2, 2019.

\bibitem{boland2019applied}
Mary~Regina Boland, Margret~L Casal, Marc~S Kraus, and Anna~R Gelzer.
\newblock Applied veterinary informatics: development of a semantic and
  domain-specific method to construct a canine data repository.
\newblock {\em Scientific reports}, 9(1):1--13, 2019.

\bibitem{inacio_2020_automated}
Sandra~Val{\'e}ria In{\'a}cio, Jancarlo Ferreira~Gomes, Alexandre
  Xavier~Falc{\~a}o, Celso~Tetsuo Nagase~Suzuki, Walter Bertequini~Nagata,
  Saulo~Hudson Nery~Loiola, Bianca Martins~dos Santos, Felipe~Augusto Soares,
  Stefani~Laryssa Rosa, Carolina~Beatriz Baptista, et~al.
\newblock Automated diagnosis of canine gastrointestinal parasites using image
  analysis.
\newblock {\em Pathogens}, 9(2):139, 2020.

\bibitem{gong_2021}
Fan Gong, Meng Wang, Haofen Wang, Sen Wang, and Mengyue Liu.
\newblock {SMR:} medical knowledge graph embedding for safe medicine
  recommendation.
\newblock {\em Big Data Research}, 23:100174, 2021.

\bibitem{gao_2021_disease}
Meng Gao, Haodong Wang, Weizheng Shen, Zhongbin Su, Huihuan Liu, Yanling Yin,
  Yonggen Zhang, and Yi~Zhang.
\newblock Disease diagnosis of dairy cow by deep learning based on knowledge
  graph and transfer learning.
\newblock {\em International Journal Bioautomation}, 25(1):87, 2021.

\bibitem{paynter_2021_veterinary}
Ashley~N Paynter, Matthew~D Dunbar, Kate~E Creevy, and Audrey Ruple.
\newblock Veterinary big data: when data goes to the dogs.
\newblock {\em Animals}, 11(7):1872, 2021.

\bibitem{li_2022_plaguekg}
Jin Li, Jing Gao, Baiyang Feng, and Yi~Jing.
\newblock Plaguekd: a knowledge graph-based plague knowledge database.
\newblock {\em The journal of biological databases and curation}, Nov 2022.

\bibitem{murali_2023_towards}
Lino Murali, G~Gopakumar, Daleesha~M Viswanathan, and Prema Nedungadi.
\newblock Towards electronic health record-based medical knowledge graph
  construction, completion, and applications: A literature study.
\newblock {\em Journal of Biomedical Informatics}, page 104403, 2023.

\bibitem{li_2020_method}
Linfeng Li, Peng Wang, Yao Wang, Shenghui Wang, Jun Yan, Jinpeng Jiang, Buzhou
  Tang, Chengliang Wang, Yuting Liu, et~al.
\newblock A method to learn embedding of a probabilistic medical knowledge
  graph: algorithm development.
\newblock {\em JMIR medical informatics}, 8:e17645, 2020.

\bibitem{lin_2020_kgnn}
Xuan Lin, Zhe Quan, Zhi{-}Jie Wang, Tengfei Ma, and Xiangxiang Zeng.
\newblock {KGNN:} knowledge graph neural network for drug-drug interaction
  prediction.
\newblock In {\em Proceedings of the 29th International Joint Conference on
  Artificial Intelligence ({IJCAI} 2020)}, volume 380, pages 2739--2745,
  Yokohama, Japan, Jul. 11--17, 2020. ijcai.org.

\bibitem{hong_2022_lagat}
Yue Hong, Pengyu Luo, Shuting Jin, and Xiangrong Liu.
\newblock Lagat: link-aware graph attention network for drug-drug interaction
  prediction.
\newblock {\em Bioinformatics}, 38(24):5406--5412, 2022.

\bibitem{nusai2015swine}
Chutchada Nusai, Sirisak Cheechang, Somkid Chaiphech, and Goragot Thanimkan.
\newblock Swine-vet: a web-based expert system of swine disease diagnosis.
\newblock {\em Procedia computer science}, 63:366--375, 2015.

\bibitem{nugroho2017mobile}
Ariadi Nugroho et~al.
\newblock Mobile expert system using fuzzy tsukamoto for diagnosing cattle
  disease.
\newblock {\em Procedia computer science}, 116:27--36, 2017.

\bibitem{DBLP:journals/access/Chai20}
Xuqing Chai.
\newblock Diagnosis method of thyroid disease combining knowledge graph and
  deep learning.
\newblock {\em {IEEE} Access}, 8:149787--149795, 2020.

\bibitem{DBLP:journals/bib/LanDCZLPC22}
Wei Lan, Yi~Dong, Qingfeng Chen, Ruiqing Zheng, Jin Liu, Yi~Pan, and
  Yi{-}Ping~Phoebe Chen.
\newblock {KGANCDA:} predicting circrna-disease associations based on knowledge
  graph attention network.
\newblock {\em Briefings Bioinform.}, 23(1), 2022.

\bibitem{DBLP:conf/icdm/XuXSLLXW21}
Xiao Xu, Xian Xu, Yuyao Sun, Xiaoshuang Liu, Xiang Li, Guotong Xie, and Fei
  Wang.
\newblock Predictive modeling of clinical events with mutual enhancement
  between longitudinal patient records and medical knowledge graph.
\newblock In {\em Proceedings of the 21st International Conference on Data
  Mining ({ICDM} 2021)}, pages 777--786, Auckland, New Zealand, Dec. 7--10,
  2021. {IEEE}.

\bibitem{lipton2015learning}
Zachary~C Lipton, David~C Kale, Charles Elkan, and Randall Wetzel.
\newblock Learning to diagnose with lstm recurrent neural networks.
\newblock arXiv preprint, arXiv:1511.03677, 2015.

\bibitem{hoangugt}
Van~Thuy Hoang and O-Joun Lee.
\newblock Transitivity-preserving graph representation learning for bridging
  local connectivity and role-based similarity.
\newblock arXiv preprint, arXiv:2308.09517, 2023.

\bibitem{kristiadi_2019}
Agustinus Kristiadi, Mohammad~Asif Khan, Denis Lukovnikov, Jens Lehmann, and
  Asja Fischer.
\newblock Incorporating literals into knowledge graph embeddings.
\newblock In {\em Proceedings of the 18th International Semantic Web Conference
  (ISWC 2019)}, volume 11778 of {\em Lecture Notes in Computer Science}, pages
  347--363, Auckland, New Zealand, Oct. 26--30, 2019. Springer.

\bibitem{chen_2020}
Ming Chen, Zhewei Wei, Zengfeng Huang, Bolin Ding, and Yaliang Li.
\newblock Simple and deep graph convolutional networks.
\newblock In {\em Proceedings of the 37th International Conference on Machine
  Learning (ICML 2020}, volume 119 of {\em Proceedings of Machine Learning
  Research}, pages 1725--1735, Virtual Event, Jul. 13--18, 2020. {PMLR}.

\bibitem{s23084168}
Van~Thuy Hoang , Hyeon-Ju Jeon , Eun-Soon You , Yoewon Yoon , Sungyeop
  Jung , and O-Joun Lee .
\newblock Graph representation learning and its applications: A survey.
\newblock {\em Sensors}, 23(8), 2023.

\bibitem{velivckovic_2017}
Petar Veli{\v{c}}kovi{\'{c}}, Guillem Cucurull, Arantxa Casanova, Adriana
  Romero, Pietro Li{\`{o}}, and Yoshua Bengio.
\newblock {Graph Attention Networks}.
\newblock {\em International Conference on Learning Representations}, 2018.

\bibitem{jeon2021learning}
Hyeon-Ju Jeon, Gyu-Sik Choi, Se-Young Cho, Hanbin Lee, Hee~Yeon Ko, Jason~J
  Jung, O-Joun Lee, and Myeong-Yeon Yi.
\newblock Learning contextual representations of citations via graph
  transformer.
\newblock In {\em Proceeding of the 2nd International Conference on
  Human-centered Artificial Intelligence (Computing4Human 2021)}, Da Nang,
  Vietnam, Oct. 2021.

\bibitem{nguyen2023connector}
Thanh~Sang Nguyen, Jooho Lee, Van~Thuy Hoang, and O-Joun Lee.
\newblock Connector 0.5: A unified framework for graph representation learning.
\newblock arXiv preprint, arXiv:2304.13195, 2023.

\bibitem{Jeon2022}
Hyeon-Ju Jeon, Min-Woo Choi, and O-Joun Lee.
\newblock Day-ahead hourly solar irradiance forecasting based on
  multi-attributed spatio-temporal graph convolutional network.
\newblock {\em Sensors}, 22(19):7179, Sep. 2022.

\bibitem{zhang_2020_graphbert}
Jiawei Zhang, Haopeng Zhang, Congying Xia, and Li~Sun.
\newblock Graph-bert: Only attention is needed for learning graph
  representations.
\newblock arXiv preprint, arXiv:2001.05140, 2020.

\bibitem{joo_2023_classification}
Yunsang Joo, Hyun{-}Cheol Park, O{-}Joun Lee, Changhan Yoon, Moon~Hyung Choi,
  and Chang Choi.
\newblock Classification of liver fibrosis from heterogeneous ultrasound image.
\newblock {\em {IEEE} Access}, 11:9920--9930, 2023.

\bibitem{zhao_2018_emr}
Chao Zhao, Jingchi Jiang, Yi~Guan, Xitong Guo, and Bin He.
\newblock Emr-based medical knowledge representation and inference via markov
  random fields and distributed representation learning.
\newblock {\em Artificial intelligence in medicine}, 87:49--59, 2018.

\bibitem{DBLP:conf/ijcai/LeeJ20}
O-Joun Lee and Jason~J. Jung.
\newblock Story embedding: Learning distributed representations of stories
  based on character networks (extended abstract).
\newblock In {\em Proceedings of the 29th International Joint Conference on
  Artificial Intelligence ({IJCAI} 2020)}, pages 5070--5074. ijcai.org, Sep.
  05, 2020.

\bibitem{DBLP:journals/ai/LeeJ20}
O{-}Joun Lee and Jason~J. Jung.
\newblock Story embedding: Learning distributed representations of stories
  based on character networks.
\newblock {\em Artificial Intelligence}, 281:103235, 2020.

\bibitem{DBLP:conf/ijcai/ParkK20}
O{-}Joun Lee, Sung~Youn Park, and Jin{-}Taek Kim.
\newblock Ideanet: Potential opportunity discovery for business innovation.
\newblock In {\em Proceedings of AI4Narratives - Workshop on Artificial
  Intelligence for Narratives in conjunction}, volume 2794 of {\em {CEUR}
  Workshop Proceedings}, pages 5--8, Yokohama, Japan, Jan. 07--08, 2020.
  CEUR-WS.org.

\bibitem{bordes_2013}
Antoine Bordes, Nicolas Usunier, Alberto Garc{\'{\i}}a{-}Dur{\'{a}}n, Jason
  Weston, and Oksana Yakhnenko.
\newblock Translating embeddings for modeling multi-relational data.
\newblock In {\em Proceedings of the 27th Annual Conference on Neural
  Information Processing Systems (NeurIPS 2013)}, volume~26, pages 2787--2795,
  Lake Tahoe, Nevada, United States, Dec. 05--08, 2013.

\bibitem{lin_2015}
Yankai Lin, Zhiyuan Liu, Maosong Sun, Yang Liu, and Xuan Zhu.
\newblock Learning entity and relation embeddings for knowledge graph
  completion.
\newblock In {\em Proceedings of the 29th Conference on Artificial Intelligence
  (AAAI 2015)}, pages 2181--2187, Austin, Texas, {USA}, Jan. 25--30 2015.
  {AAAI} Press.

\bibitem{wang_2014_knowledge}
Zhen Wang, Jianwen Zhang, Jianlin Feng, and Zheng Chen.
\newblock Knowledge graph embedding by translating on hyperplanes.
\newblock In {\em Proceedings of the 28th {AAAI} Conference on Artificial
  Intelligence (AAAI 2014)}, volume~28, pages 1112--1119, Qu{\'{e}}bec City,
  Qu{\'{e}}bec, Canada, Jul. 27--31, 2014. {AAAI} Press.

\bibitem{ji_2015_knowledge}
Guoliang Ji, Shizhu He, Liheng Xu, Kang Liu, and Jun Zhao.
\newblock Knowledge graph embedding via dynamic mapping matrix.
\newblock In {\em Proceedings of the 53rd Annual Meeting of the Association for
  Computational Linguistics ({ACL} 2015)}, pages 687--696, Beijing, China, Jul.
  26--31, 2015. ACL.

\bibitem{ji_2016_knowledge}
Guoliang Ji, Kang Liu, Shizhu He, and Jun Zhao.
\newblock Knowledge graph completion with adaptive sparse transfer matrix.
\newblock In {\em Proceedings of the 30th Conference on Artificial Intelligence
  (AAAI 2016)}, pages 985--991, Phoenix, Arizona, {USA}, Feb. 12--17, 2016.
  {AAAI} Press.

\bibitem{dettmers_2018_conv2DKG}
Tim Dettmers, Pasquale Minervini, Pontus Stenetorp, and Sebastian Riedel.
\newblock Convolutional 2d knowledge graph embeddings.
\newblock In {\em Proceedings of the 32nd Conference on Artificial Intelligence
  (AAAI 2018)}, pages 1811--1818, New Orleans, Louisiana, USA, Feb. 2--7, 2018.
  {AAAI} Press.

\bibitem{DBLP:journals/fdata/JeonLJ19}
Hyeon{-}Ju Jeon, O{-}Joun Lee, and Jason~J. Jung.
\newblock Is performance of scholars correlated to their research collaboration
  patterns?
\newblock {\em Frontiers Big Data}, 2:39, 2019.

\bibitem{tay_2017_multi}
Yi~Tay, Luu~Anh Tuan, Minh~C. Phan, and Siu~Cheung Hui.
\newblock Multi-task neural network for non-discrete attribute prediction in
  knowledge graphs.
\newblock In {\em Proceedings of the 2017 on Conference on Information and
  Knowledge Management ({CIKM} 2017)}, pages 1029--1038, Singapore, Nov.
  06--10, 2017. {ACM}.

\bibitem{wu_2018_knowledge}
Yanrong Wu and Zhichun Wang.
\newblock Knowledge graph embedding with numeric attributes of entities.
\newblock In {\em Proceedings of the 3rd Workshop on Representation Learning
  for NLP(Rep4NLP@ACL 2018)}, pages 132--136, Melbourne, Australia, Jul. 20,
  2018. ACL.

\bibitem{li_2022_fusing}
Yancong Li, Xiaoming Zhang, Fang Wang, Bo~Zhang, and Feiran Huang.
\newblock Fusing visual and textual content for knowledge graph embedding via
  dual-track model.
\newblock {\em Applied Soft Computing}, 128:109524, 2022.

\bibitem{yanaeva_2021_application}
Marina~V Yanaeva, Elena~V Kuzminova, Nikita~A Akindinov, Denis~V Osepchuk, and
  Marina~P Semenenko.
\newblock Application of intelligent methods for diagnosing animal diseases
  based on determining the structures of blood facies.
\newblock In {\em E3S Web of Conferences}, volume 262, page 02005, 2021.

\bibitem{yamaguchi_2020_deep}
Takeshi Yamaguchi, Kenichi Inoue, Hiroko Tsunoda, Takayoshi Uematsu, Norimitsu
  Shinohara, and Hirofumi Mukai.
\newblock A deep learning-based automated diagnostic system for classifying
  mammographic lesions.
\newblock {\em Medicine}, 99(27), 2020.

\bibitem{gesese_2021_survey}
Genet~Asefa Gesese, Russa Biswas, Mehwish Alam, and Harald Sack.
\newblock A survey on knowledge graph embeddings with literals: Which model
  links better literally?
\newblock {\em Semantic Web}, 12(4):617--647, 2021.

\bibitem{dai_2020_survey}
Yuanfei Dai, Shiping Wang, Neal~N Xiong, and Wenzhong Guo.
\newblock A survey on knowledge graph embedding: Approaches, applications and
  benchmarks.
\newblock {\em Electronics}, 9(5):750, 2020.

\bibitem{shang_2021_ehr}
Yong Shang, Yu~Tian, Min Zhou, Tianshu Zhou, Kewei Lyu, Zhixiao Wang, Ran Xin,
  Tingbo Liang, Shiqiang Zhu, and Jingsong Li.
\newblock Ehr-oriented knowledge graph system: Toward efficient utilization of
  non-used information buried in routine clinical practice.
\newblock {\em {IEEE} Journal of Biomedical and Health Informatics},
  25(7):2463--2475, 2021.

\bibitem{bojanowski_2017_fasttext}
Piotr Bojanowski, Edouard Grave, Armand Joulin, and Tom{\'{a}}s Mikolov.
\newblock Enriching word vectors with subword information.
\newblock {\em Transactions of the association for computational linguistics},
  5:135--146, 2017.

\bibitem{wang_2019}
Xiang Wang, Xiangnan He, Yixin Cao, Meng Liu, and Tat{-}Seng Chua.
\newblock {KGAT:} knowledge graph attention network for recommendation.
\newblock In {\em Proceedings of the 25th {ACM} {SIGKDD} International
  Conference on Knowledge Discovery {\&} Data Mining ({KDD} 2019)}, pages
  950--958, Anchorage, AK, USA, Aug. 04--08, 2019. {ACM}.

\bibitem{hamilton_2017}
William~L. Hamilton, Zhitao Ying, and Jure Leskovec.
\newblock Inductive representation learning on large graphs.
\newblock In {\em Proceedings of the 30th Annual Conference on Neural
  Information Processing Systems (NeurIPS 2017)}, pages 1024--1034, Long Beach,
  CA, {USA}, Dec. 04--09, 2017.

\bibitem{xu_2018}
Keyulu Xu, Weihua Hu, Jure Leskovec, and Stefanie Jegelka.
\newblock How powerful are graph neural networks?
\newblock In {\em Proceedings of the 7th International Conference on Learning
  Representations ({ICLR} 2019)}, New Orleans, LA, USA, May 06--09, 2019.
  OpenReview.net.

\bibitem{tang_2015_line}
Jian Tang, Meng Qu, Mingzhe Wang, Ming Zhang, Jun Yan, and Qiaozhu Mei.
\newblock {LINE:} large-scale information network embedding.
\newblock In {\em Proceedings of the 24th International Conference on World
  Wide Web ({WWW} 2015)}, pages 1067--1077, Florence, Italy, May 18--22, 2015.
  {ACM}.

\bibitem{jiang2022kgnmda}
Changzhi Jiang, Minli Tang, Shuting Jin, Wei Huang, and Xiangrong Liu.
\newblock {KGNMDA:} {A} knowledge graph neural network method for predicting
  microbe-disease associations.
\newblock {\em {IEEE} {ACM} Transactions on Computational Biology and
  Bioinformatics}, 20(2):1147--1155, 2023.

\bibitem{kingma_2014_adam}
Diederik~P. Kingma and Jimmy Ba.
\newblock Adam: {A} method for stochastic optimization.
\newblock In {\em Proceedings of the 3rd International Conference on Learning
  Representations ({ICLR} 2015)}, San Diego, CA, USA, May 07--09, 2015.

\bibitem{kipf_2016}
Thomas~N. Kipf and Max Welling.
\newblock Semi-supervised classification with graph convolutional networks.
\newblock In {\em 5th International Conference on Learning Representations
  ({ICLR} 2017)}, Toulon, France, Apr. 24--26, 2017. OpenReview.net.

\end{thebibliography}

\end{document}